\documentclass{article} 
\usepackage{iclr2027_conference,times}

\usepackage{amsmath,amsfonts,bm}

\def\eqref#1{equation~\ref{#1}}

\def\1{\bm{1}}

\DeclareMathAlphabet{\mathsfit}{\encodingdefault}{\sfdefault}{m}{sl}
\SetMathAlphabet{\mathsfit}{bold}{\encodingdefault}{\sfdefault}{bx}{n}

\usepackage{hyperref}
\usepackage{url}
\usepackage[dvipsnames]{xcolor}
\usepackage{booktabs}
\usepackage{tabularx}
\usepackage{graphicx}
\usepackage{subcaption}
\usepackage[capitalize]{cleveref}
\usepackage{csquotes}
\usepackage{wrapfig}
\usepackage{multirow}
\usepackage{tcolorbox}
\tcbuselibrary{breakable, listings}
\usepackage{comment}

\usepackage[T1]{fontenc}
\usepackage[varqu,scaled=0.95]{zi4}   
\usepackage[dvipsnames]{xcolor}       
\usepackage{tcolorbox}
\tcbuselibrary{listings,breakable,skins}

\newcommand{\promptlisting}[2]{%
  \tcbinputlisting{
    enhanced,
    breakable,
    listing only,
    listing file={#2},
    title={#1},
    colback=gray!3,
    colframe=Blue!70!black,
    colbacktitle=Blue!12,
    coltitle=Blue!70!black,
    fonttitle=\bfseries\small,
    boxrule=0.5pt,
    arc=2pt,
    left=6pt, right=6pt, top=4pt, bottom=4pt,
    toptitle=2pt, bottomtitle=2pt,
    listing options={
      basicstyle=\ttfamily\footnotesize,
      breaklines=true,
      breakatwhitespace=true,
      columns=fullflexible,
      keepspaces=true,
      showstringspaces=false,
      breakindent=0pt,
    },
  }%
}

\title{Agents are systems, not models: \\ Rethinking agentic evaluation}

\author{Luis Wiedmann\textsuperscript{1,2}, Leander Girrbach\textsuperscript{1,2}, Cordelia Schmid\textsuperscript{3}, Zeynep Akata\textsuperscript{1,2} \\
\textsuperscript{1}Technical University of Munich, MCML \hspace{0.5cm}
\textsuperscript{2}Helmholtz Munich \\
\textsuperscript{3}Inria, École normale supérieure, CNRS, PSL Research University
}

\iclrfinalcopy 
\begin{document}

\maketitle

\begin{abstract}
Agent evaluations increasingly go beyond a single success rate, reporting metrics such as cost, consistency, and robustness. Yet they typically treat the agent itself as fixed. In practice, an agent is a configurable system: users decide what to tell it, how long to let it run, and which model to use, and each of these choices can change how well and how consistently it performs. 
To study these choices, we build a testbed of four scientific tasks in which a coding agent must find and correctly operate a published specialist model. 
We vary five parts of the agent's configuration in a full factorial design: task information, reasoning, self-verification, time budget, and backbone model.
We find that a single run is not a reliable measure of a configuration: rerunning the same configuration changes whether the agent even finishes in 42\% of configurations, and about five repeated runs are needed to compare configurations reliably.
Across our main experiments, the information provided to the agent has the largest effect, exceeding both time budget and model size, while also reducing cost and improving calibration.
Configuration choices also interact: additional time helps only when the agent has sufficient information or a capable enough model to use it.
Finally, a trajectory-based taxonomy of agent behavior reveals that prompting an agent to verify its answer has little effect on its verification behavior, whereas providing a dedicated verification tool changes that behavior substantially.
These results suggest that agents should be evaluated as configurable systems, and that some desired behaviors are more effectively enabled at the system level than requested through prompting. We release the benchmark and more than 18{,}000 agent trajectories\footnote{\url{https://github.com/lusxvr/rethinking-agent-evaluation}}.
\end{abstract}
\section{Introduction}
\label{sec:intro}

When agents are used to solve complex tasks, we care not only about whether they succeed, but also about how reliably they do so, and about which parts of their configuration determine capability and reliability. Recent work broadens what is measured about an agent: \citet{kapoor2025ai} argue that evaluations should not focus only on performance, but also account for cost and comparability between methods. \citet{rabanser2026towards} show that capability gains do not automatically translate into reliability gains, and argue for a decomposition of agent reliability into consistency, robustness, predictability, and safety. 
It remains less clear how an agent's configuration itself shapes these measurements: which of its parts drive capability and reliability, and how these parts interact.

Scientific workflows are a natural setting to study this. Many scientific tasks rely on specialist models (trained for a specific scientific domain or task) that a general-purpose LLM cannot directly replace, so an agent cannot solve them from its own knowledge. Instead, it has to find the right model, prepare the input data, and write the code to run it, and each of these steps can be affected by how the agent is configured. Published results of these specialist models further provide reference scores against which the agent's outcome can be checked. One example is estimating redshift from galaxy images. Fine-tuning a multimodal LLM to process galaxy images would be costly, but specialist models for this task already exist, and an agent can use them instead. We are interested in \emph{how reliably agents can use such specialist models} in scientific workflows, and \emph{what contributes to their effective use}.

In this paper, we use the following terms: the \textit{model} (or \textit{backbone}) is the LLM itself, with no memory across calls and no ability to act on its own. An \textit{agent} wraps a model in a loop that lets it act repeatedly: read the current state, decide on an action, execute it, observe the result, and decide again. The agent's \textit{harness} is the infrastructure that runs this loop: it dispatches tool calls and manages the model's context at each step, including memory and compression. Fundamentally, an agent is a \textit{system} consisting of a model and a harness \citep{de2026makes, paniego2026glossary}, which is reflected in recent implementations of production coding agents such as Codex or Claude Code \citep{lopopolo2026harness, young2025harnesses, shi2026programming}. 

We systematically evaluate how an agent's configuration affects how well it uses specialist models to solve scientific tasks. Concretely, we study five axes of the agent's configuration: (1) what \textit{information} the agent is given; (2) whether the agent generates \textit{reasoning} and keeps it across steps; (3) how strongly the agent is asked to \textit{verify} its own answer; (4) how much runtime \textit{budget} the agent has to solve the task; and (5) how strong the agent's backbone \textit{model} is. 
Because users set much of this configuration each time they use an agent, a score measured in one configuration does not necessarily predict the outcome for another user. We therefore vary all five axes in a full factorial design with five repeated runs per configuration. This design measures how each axis contributes to the agent's outcome and reliability, and how the axes interact, which one-factor-at-a-time ablations cannot reveal. As a testbed, we build a benchmark of four scientific tasks from astrophysics and genomics, in which the agent's outcome is compared against the specialist model on an evaluation from its original paper.

We find that a single run is not a reliable measure of a configuration: rerunning the same configuration often changes whether the agent finishes at all, and about five repeated runs per configuration are needed to compare configurations reliably. 
In our main experiments, the information given to the agent has the largest effect among the axes, outranking both the budget and the backbone model. The effect of one axis can also depend on another: additional budget mainly improves the score when the agent already has enough information to use it productively. Finally, an analysis of the agent's behavior shows that some behaviors, such as self-verification, are hard for the user to influence directly through prompting and should instead be shaped by the system's design.

In summary, our contributions are as follows: \textbf{(1)} We systematically investigate how the axes of an agent's configuration influence its outcome and reliability through a full factorial study. It shows that single runs do not reliably predict a configuration's performance, that information has the largest effect in our main experiments, and that axes interact strongly enough to reverse the direction of an effect. \textbf{(2)} We create an AI-for-Science (AI4S) benchmark in which the agent has to use specialist models under controlled conditions. We release the benchmark and 18{,}240 agent trajectories with their produced artifacts, enabling reproducibility and further analysis of agent behavior. \textbf{(3)} We derive a taxonomy of agent behavior from these trajectories and use it to show that some behaviors are hard for the user to influence directly and should instead be shaped by the system's design.

\section{Related Work}
\label{sec:related-work}

\paragraph{AI for Science.}
Recent capability gains in machine learning methods and foundation models enable a broad range of applications in scientific fields outside classical computer science \citep{jumper2021highly, wang2023scientific}. Furthermore, the transition from isolated models to agents has intensified the interest in AI as a scientific companion or as a scientist itself \citep{wei2025ai, gridach2025agentic, xin2025towards, lu2026towards}. We organize the field into two focus areas: \emph{(re-)implementation}~\citep{starace2025paperbench} and \emph{discovery}~\citep{majumder2025discoverybench}. Works on (re-)implementation reuse scientific publications or their codebase, and test how well agents can reproduce the findings, methods, or implementations \citep{wang2026naturebench, chen2025scienceagentbench, kim2026reproduction, bogin2024super, falck2026training, shen2026sciagentgym}. In parallel, a growing set of works evaluates agents' ability to produce genuinely novel discoveries \citep{kirgis2026can, artiles2026alien, si2025can, zahavy2026position, si2026the, messeri2024artificial}. Our work is closer to the first area, i.e., (re-)implementation of existing work. We propose a coding-focused benchmark that enables the agent to freely explore, test, and write code to operate given specialist models, while its outcome is judged against their published performance. 
This makes the environment controlled and repeatable, so that the agent's configuration can be varied systematically and outcomes can be scored without manual verification.

\vspace{-0.05cm}
\paragraph{Agents and Their Evaluation.}
As agents~\citep{wooldridge1995intelligent, yao2023react} become more capable, evaluating their performance~\citep{chang2024survey} becomes more challenging. Their ability to tackle complex, long-horizon tasks~\citep{kwa2025measuring}, potentially through multiple instances~\citep{cemri2026why}, makes failures and capabilities harder to accurately estimate and understand, which makes a systematic study necessary~\citep{zhu2026establishing, miller2024adding}. Multiple works agree that a single success metric is insufficient, and that capabilities have to be judged both in the context of their task and in terms of their cost and reliability, broadening what is measured about an agent~\citep{liang2023holistic, kapoor2026holistic, kapoor2025ai, rabanser2026towards}. 
\citet{zhuge2025agentasajudge} argue that judging only an agent's final output, or relying on human evaluation, is insufficient, and use an agent to check intermediate requirements on tasks that lack an automatic score. \citet{zhu2025llm} and \citet{wang2026deep} locate where agents fail within their trajectories. These works improve how outcomes are measured, but each evaluates agents in a fixed setup. We complement them by treating the agent as a configurable system rather than a fixed model, and study how its configuration shapes these measurements.

\section{Agent Evaluation and Benchmark Tasks}
\label{sec:method}

\begin{figure}[t]
    \centering
    \includegraphics[width=\linewidth]{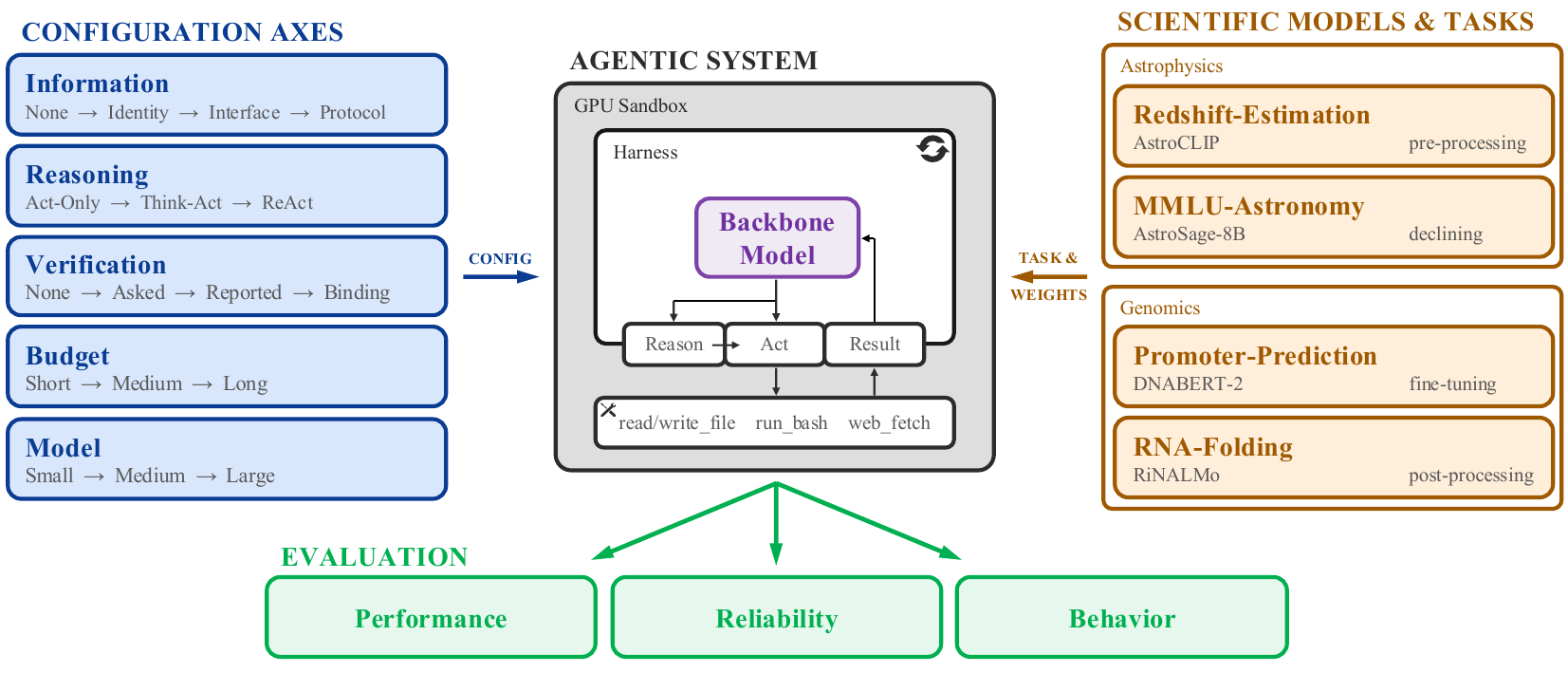}
    \caption{Architecture of our agentic AI4S benchmark. The agent (center) has access to different specialist models and is evaluated against their published performance (right). By varying the axes of the agent's configuration within the same task (left), we can systematically study its performance, reliability, and behavior under controlled, repeatable conditions (bottom).}
    \label{fig:architecture}
\end{figure}

In our setting, a single agent has access to a collection of specialist models, including their weights and pre-built environments (i.e., the libraries and dependencies required to run them). Given a task, the agent has to determine which model to use (if any) and then correctly operate it by writing its own code. In the following, we describe how our agents and tasks are implemented (see \cref{fig:architecture} for an overview).

\subsection{Modular Agent Implementation for Scientific Workflows}
The agent runs in a sandbox with access to a single GPU and a pre-built coding environment. The sandbox contains the weights and environments of all specialist models, each with a short README that names the model and links to its paper and code. 
Importantly, the specialist models are not available through a dedicated tool call, but have to be operated by the agent the same way a human scientist would. The agent receives a simple system prompt that describes it as an \enquote{autonomous research agent} (\cref{sec:apx-agent-prompts}). It is then given the task description, the available data, and the required solution format. It can read and write files, fetch web pages, and run terminal commands, and can therefore write and execute its own code.

In each step, the agent's backbone model sees the conversation so far, including the task description, all outputs, and every tool result. Then, it either outputs plain text or calls one of the available tools (\cref{sec:apx-tool-use}). A tool's output is appended to the conversation before the loop repeats, so the agent always sees the outcome of its last action before choosing its next one. The loop ends when the agent calls a \texttt{finish} tool with its answer, or when the budget runs out. In its full form, this loop follows \citet{yao2023react}, interleaving reasoning about what to do with the actions taken.

For our main analysis, we select and systematically vary five axes, each targeting a distinct factor known to shape behavior. What agents are told determines what they still need to explore or retrieve themselves \citep{jimenez2024swebench}. Keeping reasoning in context is central to the performance gains of \citet{yao2023react}, and agents that check their own work tend to succeed more often \citep{shinn2023reflexion,madaan2023self}. Giving agents more time lets them solve harder problems, though with diminishing returns \citep{wijk2025rebench, chan2025mle}. Finally, the capability of the backbone model should be evaluated separately from the harness built around it \citep{kapoor2025ai}. Together, the five axes cover what the agent knows, how it deliberates, whether it checks itself, how much time it has, and how capable it is, all of which can influence both outcome and reliability. We describe the axes and their levels below. For \textit{Information} and \textit{Verification}, each level adds to the previous one (marked with +).

\textbf{Information.} This axis controls how much information the agent receives in addition to the task description: (i) \emph{none}: no additional information; (ii) \emph{identity}: which specialist model is associated with the task; (iii) \emph{interface}: + how to load and call the model; (iv) \emph{protocol}: + a full recipe for this specific task.

\textbf{Reasoning.} This axis controls whether the agent generates reasoning and whether this reasoning stays in context for later steps: (i) \emph{act-only}: no reasoning is generated; (ii) \emph{think-act}: reasoning is generated in each step but discarded before the next; (iii) \emph{ReAct}: reasoning is generated and kept in context across steps.

\textbf{Verification.} This axis controls how strongly the prompt asks the agent to verify its own answer: (i) \emph{none}: no instruction to verify; (ii) \emph{asked}: the agent is asked to verify its answer; (iii) \emph{reported}: + the agent must state the score it expects and the evidence from its verification; (iv) \emph{binding}: + the agent is told not to submit unless its own verification is convincing.

\textbf{Budget.} This axis controls how much wall-clock time the agent has to solve the task: (i) \emph{short}: 5 min; (ii) \emph{medium}: 10 min; (iii) \emph{long}: 20 min.
We limit runtime rather than the number of steps, because the duration of a step varies widely, from a quick file read to fine-tuning a specialist model. We also do not limit tokens or token cost, because such a limit would constrain only generation and not the execution of code, which takes most of a run's time (\cref{sec:apx-speed}).

\textbf{Model.} This axis controls the size, and therefore the capability, of the agent's backbone model: (i) \emph{small}: Qwen3.5-35B-A3B; (ii) \emph{medium}: Qwen3.5-122B-A10B; (iii) \emph{large}: Qwen3.5-397B-A17B. All models are FP8-quantized due to resource limitations.

\subsection{Benchmark Tasks from Astrophysics and Genomics}
To create the tasks, we identify suitable published scientific works that propose a new specialist model. We select these works from 255 candidate papers found by a systematic literature search (\cref{sec:apx-task-selection}). For each, we reproduce the evaluation from the original publication ourselves, reducing its size where necessary so that it can be completed within the shortest budget (e.g., by downsampling the evaluation dataset). The score of the specialist model in our reproduction is the task's reference score. Each task tests, in a controlled setup, the agent's ability to understand, reproduce, and use prior work to complete a scientific workflow. We include the following tasks (\cref{tab:tasks-and-examples,sec:apx-task-examples}):

\textbf{\texttt{redshift-estimation}.} The agent has to estimate the cosmological redshift of galaxies directly from multi-band photometric images. Given 20 galaxy image cutouts and a labeled reference set, it has to use the pretrained AstroCLIP model \citep{parker2024astroclip} to predict continuous redshift values. Performance is measured by $R^2$ against spectroscopic ground-truth measurements.

\textbf{\texttt{mmlu-astronomy}.} This task consists of 152 multiple-choice questions on topics from stellar evolution to cosmology. The specialist model, AstroSage-8B \citep{de2025astrosage}, is weaker than the agent's backbone model, so the task tests whether the agent recognizes this and answers the questions itself instead of relying on the specialist. Performance is measured by accuracy.

\textbf{\texttt{promoter-prediction}.} The agent has to classify 613 human DNA sequences as promoter or non-promoter regions. To do so, it has to locate, pre-process, fine-tune, and operate DNABERT-2 \citep{zhou2024dnabert}. Performance is measured by the Matthews correlation coefficient (MCC).

\textbf{\texttt{rna-folding}.} The agent has to predict the secondary structures of 300 RNA sequences. It has to use the pretrained RiNALMo model \citep{penic2025rinalmo} to predict contacts between nucleotide pairs, and then apply a post-processing procedure to obtain valid secondary structures. Performance is measured by the structure-level pairing $F_1$ score.

We distinguish two regimes: a task is \textsc{gap\_positive} if the specialist's reference score is higher than the score of the backbone model alone ($\mathcal{R} > \mathcal{B}$), and \textsc{gap\_negative} if it is lower. 
The four tasks each stress a different part of using a specialist model correctly.
\texttt{redshift-estimation} and \texttt{rna-folding} reward finding a stronger specialist and using it without fine-tuning, but the difficult step differs (pre-processing input images before embedding them, and post-processing the predicted contacts into valid structures). \texttt{promoter-prediction} instead requires fine-tuning DNABERT-2, because its pretrained checkpoint is not trained for promoter classification. \texttt{mmlu-astronomy} is \textsc{gap\_negative}: its specialist model has been surpassed by general-purpose LLMs and is weaker than the backbone, so the correct behavior is to decline the specialist rather than use it.
With two domains, two regimes, and four different demands on the specialist model, no single fixed strategy (always use the model, or never use it) performs well across all four tasks. To succeed, the agent has to investigate and judge each specialist's capabilities and work out its specific operating details.

\begin{table}[t]
    \centering
    \caption{The four tasks of the benchmark, together with one example instance per task. Each task belongs to one of two domains and one of two regimes, and focuses on a different aspect of the specialist models' operation. 
    }
    \label{tab:tasks-and-examples}
    \small
    \resizebox{\textwidth}{!}{%
    \begin{tabular}{@{}lp{2.7cm}p{3cm}p{3cm}p{3cm}@{}}
        \toprule
         & \texttt{redshift-} \texttt{estimation} & \texttt{mmlu-} \texttt{astronomy} & \texttt{promoter-} \texttt{prediction} & \texttt{rna-folding} \\
        \midrule
        Domain & Astrophysics & Astrophysics & Genomics & Genomics \\
        Size & 20 galaxy images & 152 questions & 613 DNA sequences & 300 RNA sequences \\
        Metric & $R^2$ & Accuracy & MCC & $F_1$ \\
        Specialist & AstroCLIP & AstroSage-8B & DNABERT-2 & RiNALMo \\
        Challenge & Pre-processing & Declining & Fine-tuning & Post-processing \\
        Regime & \textsc{gap\_positive} & \textsc{gap\_negative} & \textsc{gap\_positive} & \textsc{gap\_positive} \\
        \bottomrule
    \end{tabular}%
    }
    \includegraphics[width=\linewidth]{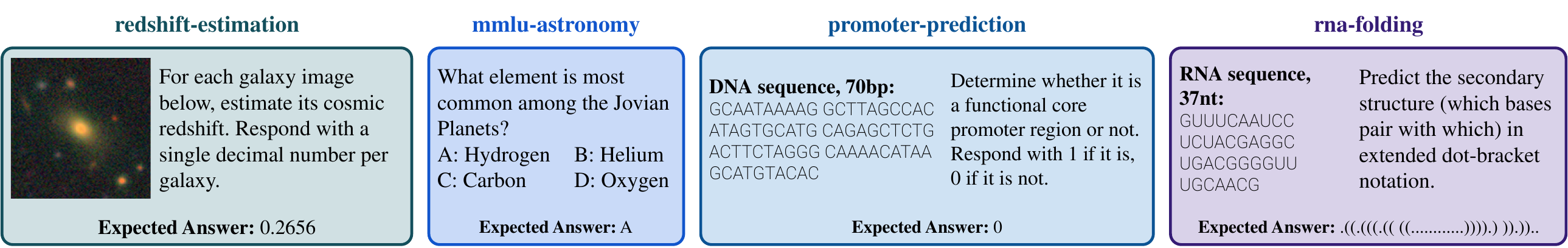}
\end{table}

\section{Experiments and Results}
\label{sec:results}

\subsection{Experimental Setup and Evaluation Metrics}
To evaluate the agent, we use three quantities. $\mathcal{B}$ is the score of the backbone model alone, i.e., the LLM answering the task directly, without an agent or specialist model. $\mathcal{R}$ is the reference score of the specialist model in our reproduction. $\mathcal{S}$ is the agent's score on the task. Our main metric is the share of the gap between backbone and specialist that the agent closes (\textsc{gap\_closed}):
\begin{equation}
    \mathcal{G} = \frac{\mathcal{S} - \mathcal{B}}{\mathcal{R} - \mathcal{B}},
\end{equation}
where $\mathcal{G}=0$ means no improvement over the backbone alone and $\mathcal{G}=1$ means the agent matches the specialist's reference score. It is defined only for \textsc{gap\_positive} tasks ($\mathcal{R} > \mathcal{B}$).
A run \emph{completes} if the agent submits an answer with the \texttt{finish} tool before its budget runs out; runs that do not complete receive no score. For every completed run, we also record a binary \emph{hurdle} indicator $\mathcal{H}$, which states whether the run's score exceeds the trivial score $\mathcal{T}$ of the task by more than a margin $\mu$ (\cref{sec:apx-margin}). $\mathcal{H}$ thus captures whether a run makes a genuine attempt at the task, and $\mathcal{G}$ captures how close it comes to the reference.
For each task, we evaluate the full grid over our axes: 4 (\emph{Information}) $\times$ 3 (\emph{Reasoning}) $\times$ 4 (\emph{Verification}) $\times$ 3 (\emph{Budget}) $\times$ 3 (\emph{Model}) $=$ 432 configurations, which we call \emph{cells}. To measure how much outcomes vary between runs, we repeat every cell 5 times, which gives 2{,}160 runs per task and $4 \times 2{,}160 = 8{,}640$ runs in total for our main experiments.

\subsection{Results}
The four tasks differ widely in both score and reliability (\cref{tab:results-overview}). \texttt{promoter-prediction} is the most reliable: almost all of its completed runs clear the hurdle, and repeated runs of the same configuration agree closely. \texttt{redshift-estimation} is by far the least reliable, with more than half of its completed runs failing to clear the hurdle. Unless noted otherwise, the remaining analyses pool the three \textsc{gap\_positive} tasks, for which $\mathcal{G}$ is defined.

\begin{table}[t]
    \centering
    \caption{Results per task for the main experiments (432 cells $\times$ 5 runs). Completion rate is computed over all runs; all other outcome rows are computed over completed runs. $C_{\text{out}}$ is computed over cells with at least 3 completed runs. Full breakdown in \cref{tab:hurdle-decomposition}. $^\dagger$Raw scale, not comparable across tasks.
    }
    \label{tab:results-overview}
    \small
    \resizebox{\textwidth}{!}{%
    \begin{tabular}{@{}lrrrr@{}}
        \toprule
         & \texttt{redshift-} & \texttt{mmlu-} & \texttt{promoter-} & \texttt{rna-} \\
         & \texttt{estimation} & \texttt{astronomy} & \texttt{prediction} & \texttt{folding} \\
        \midrule
        \multicolumn{5}{@{}l}{\textit{Runs \& Completion}} \\
        \# Runs                                  & 2{,}160  & 2{,}160  & 2{,}160  & 2{,}160  \\
        Completion Rate                         & 76.3\% $\pm$ 1.8\% & 95.3\% $\pm$ 0.9\% & 89.3\% $\pm$ 1.3\% & 61.7\% $\pm$ 2.0\% \\
        \% Cells that (always / mixed / never) complete   & 54.6 / 40.3 / 5.1 & 81.9 / 18.1 / 0.0 & 63.7 / 36.1 / 0.2 & 21.3 / 72.0 / 6.7 \\
        \midrule
        \multicolumn{5}{@{}l}{\textit{Anchors}} \\
        \textsc{Reference} $\mathcal{R}$    & 0.753  & 0.671  & 0.746  & 0.735  \\
        \textsc{Backbone} $\mathcal{B}$     & $-$2.701 & 0.967  & $-$0.030 & 0.000  \\
        \textsc{Trivial} $\mathcal{T}$      & 0.000  & 0.250  & 0.000  & 0.000  \\
        \textsc{Margin} $\mu$               & 0.250  & 0.050  & 0.100  & 0.100  \\
        \midrule
        \multicolumn{5}{@{}l}{\textit{Outcome}} \\
        Mean Score $\mathcal{S}$$^\dagger$    & $-$0.109 & 0.685 & 0.625 & 0.628 \\
        Mean \textsc{gap\_closed} $\mathcal{G}$              & 0.750 & --- & 0.844 & 0.854 \\
        Hurdle Failure Rate & 53.8\% $\pm$ 2.4\% & 4.7\% $\pm$ 0.9\% & 0.5\% $\pm$ 0.4\% & 8.7\% $\pm$ 1.5\% \\
        Outcome Consistency $C_{\text{out}}$   & 0.727 & 0.905 & 0.986 & 0.825 \\
        \bottomrule
    \end{tabular}%
    }
\end{table}

\begin{figure}[t]
    \centering
    \begin{minipage}[c]{0.7\textwidth}
        \centering
        \includegraphics[width=\textwidth]{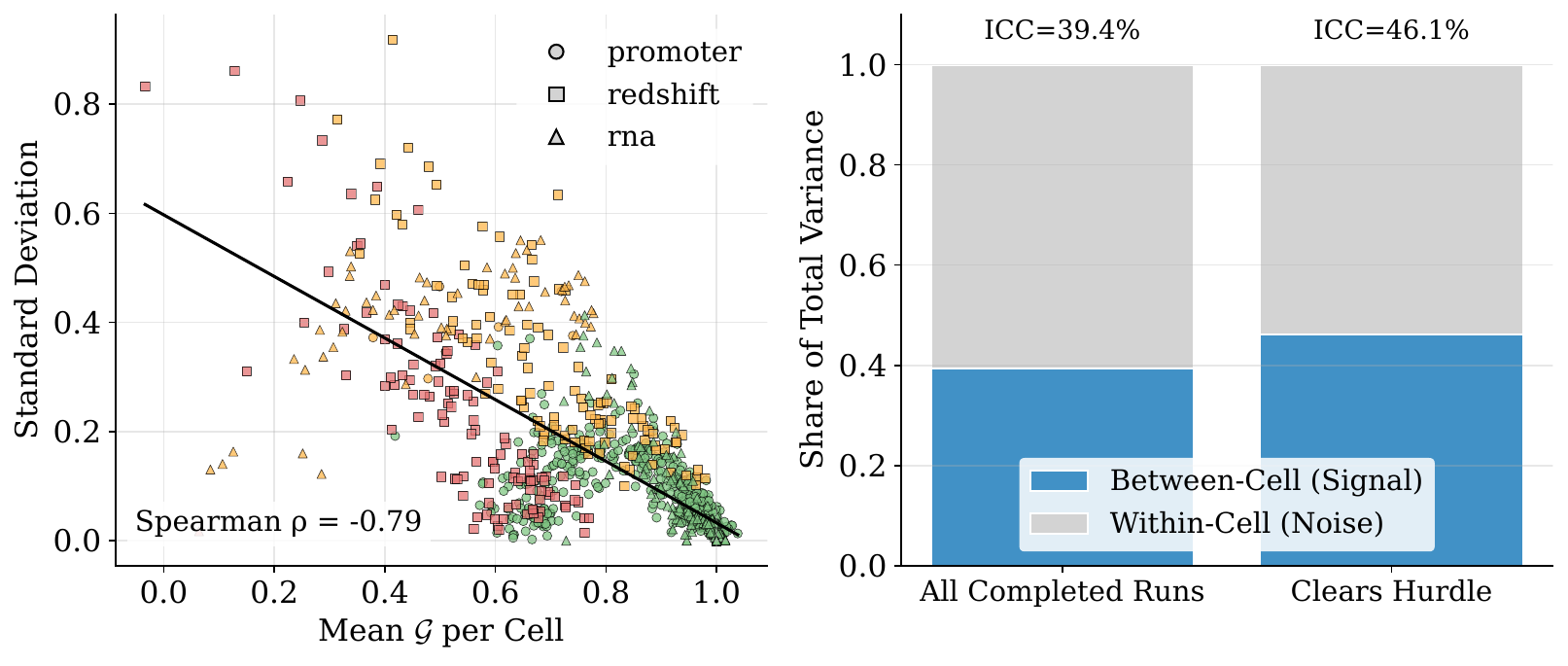}
    \end{minipage}%
    \hfill
    \begin{minipage}[c]{0.3\textwidth}
        \captionof{figure}{Outcome variance across cells. \textit{Left}: mean $\mathcal{G}$ vs.\ SD per cell ($\geq$3 completed runs), with Spearman correlation. Color: all/some/no runs clear the hurdle (green/orange/red). \textit{Right}: variance between vs.\ within cells, for all completed runs and hurdle-clearing runs only.}
        \label{fig:variance}
    \end{minipage}
\end{figure}

\paragraph{A single run is not a reliable measure of a configuration.}
Repeated runs of the same configuration often disagree on whether the agent finishes at all: across the four tasks, 42\% of cells complete in some runs but not in others, ranging from 18\% to 72\% depending on the task (\cref{tab:results-overview}). Scores vary substantially as well. Even among genuine attempts, differences between configurations explain only 46.1\% of the score variance, and the remaining 53.9\% comes from repeating the same configuration (\cref{fig:variance}, right). Without excluding runs that fail the hurdle, configurations explain only 39.4\%. The share explained by configurations is the intraclass correlation \citep[ICC;][]{shrout1979intraclass}, which is also the \emph{reliability} of a single run that can be raised by averaging over repeated runs \citep{brennan1992generalizability}. While five runs suffice on average to tell configurations apart among genuine attempts (\cref{tab:reliability}), the 95\% confidence interval of an individual cell's mean score remains wide (\cref{sec:apx-noise-floor}). The noise is also not uniform: cells with higher scores have systematically lower variance (\cref{fig:variance}, left). Finally, whether a run clears the hurdle is also inconsistent across repeated runs, and this differs strongly between tasks: the pooled outcome consistency is 0.86, but only 0.73 for \texttt{redshift-estimation} (\cref{tab:results-overview,tab:hurdle-decomposition}).
Given the level of run-to-run noise and the uneven completion rates across configurations and tasks, our results reinforce recent calls for repeated runs and for more rigorous reporting of completion rates, variance, and confidence intervals alongside raw scores as standard practice in agent evaluations \citep{kapoor2025ai, rabanser2026towards}.

\paragraph{Information has the largest axis effect.}
We rank the five axes by effect size: the range of an axis's mean $\mathcal{G}$ across its levels, divided by the within-cell standard deviation. \emph{Information} has the largest effect on every task and clearly outranks \emph{Model} (\cref{tab:axis-effects}). On average, \emph{Reasoning} matters more than \emph{Budget}, and \emph{Verification} ranks last. Because the levels of an axis complete at different rates, a ranking computed over completed runs can be affected by which runs complete. The ranking is nevertheless robust: \emph{Information} ranks first in at least 99\% of bootstrap samples, and a linear model gives the same order (\cref{sec:apx-effect-sizes}). This suggests that a user configuring an agent may benefit more from prioritizing task information than from choosing a stronger backbone model or a larger budget. The effect of one axis can, however, depend on the others, which we examine next.
\begin{table}[t]
    \centering
    \caption{Effect of each axis on $\mathcal{G}$ per task (range of mean $\mathcal{G}$ across the axis's levels, divided by the within-cell standard deviation). Axes are ranked within each task (1 = largest effect).}
    \label{tab:axis-effects}
    \small
    \begin{tabularx}{\textwidth}{@{\extracolsep{\fill}}@{}lccccccc@{}}
        \toprule
        & \multicolumn{2}{c}{\texttt{promoter-prediction}} & \multicolumn{2}{c}{\texttt{redshift-estimation}} & \multicolumn{2}{c}{\texttt{rna-folding}} & \multicolumn{1}{c}{} \\
        \cmidrule(lr){2-3} \cmidrule(lr){4-5} \cmidrule(lr){6-7}
        Axis & Rank & Effect & Rank & Effect & Rank & Effect & Mean Rank \\
        \midrule
        Information  & 1 & 1.925 & 1 & 1.452 & 1 & 1.496 & 1.00 \\
        Model        & 2 & 0.704 & 4 & 0.170 & 2 & 1.253 & 2.67 \\
        Reasoning    & 3 & 0.274 & 3 & 0.172 & 3 & 0.302 & 3.00 \\
        Budget       & 4 & 0.142 & 2 & 0.201 & 4 & 0.255 & 3.33 \\
        Verification & 5 & 0.131 & 5 & 0.130 & 5 & 0.077 & 5.00 \\
        \bottomrule
    \end{tabularx}
\end{table}

\begin{figure}[t]
  \centering
  \begin{subfigure}{0.605\textwidth}
      \centering
      \includegraphics[width=\textwidth]{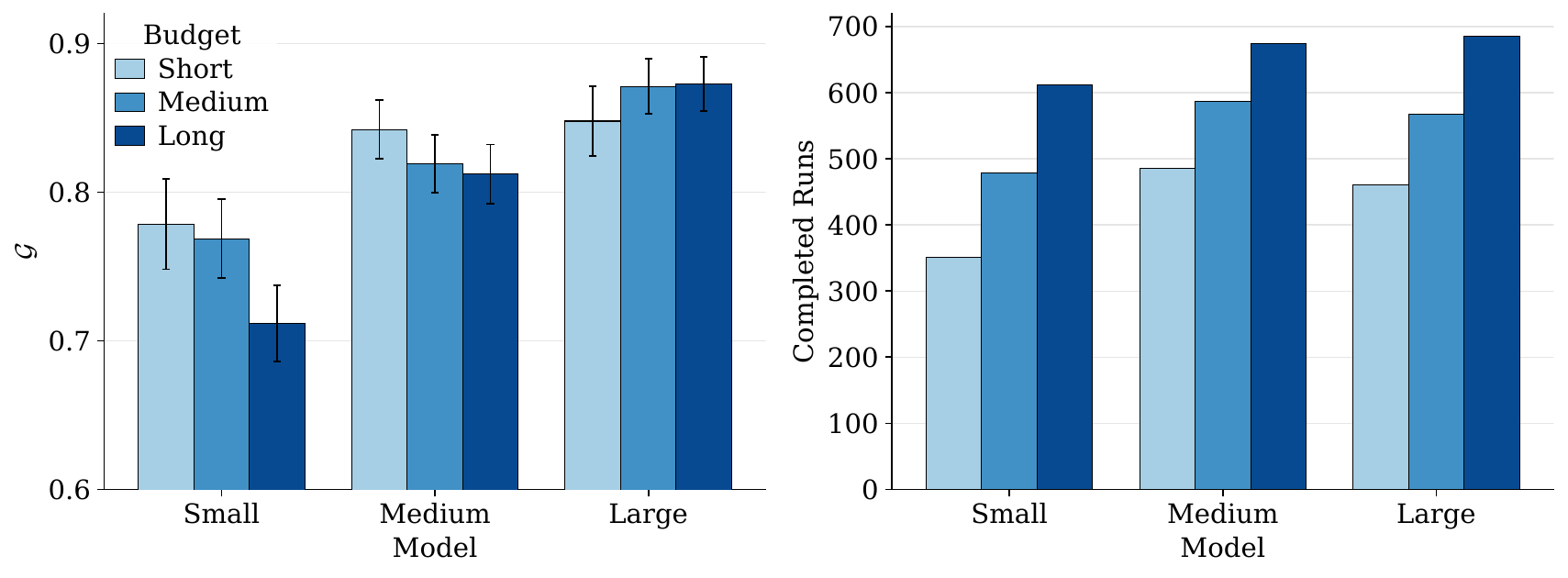}
      \caption{Interaction of the \emph{Model} and \emph{Budget} axes.}
      \label{fig:model-budget_interactions}
  \end{subfigure}
  \hfill
  \begin{subfigure}{0.38\textwidth}
      \centering
      \includegraphics[width=\textwidth]{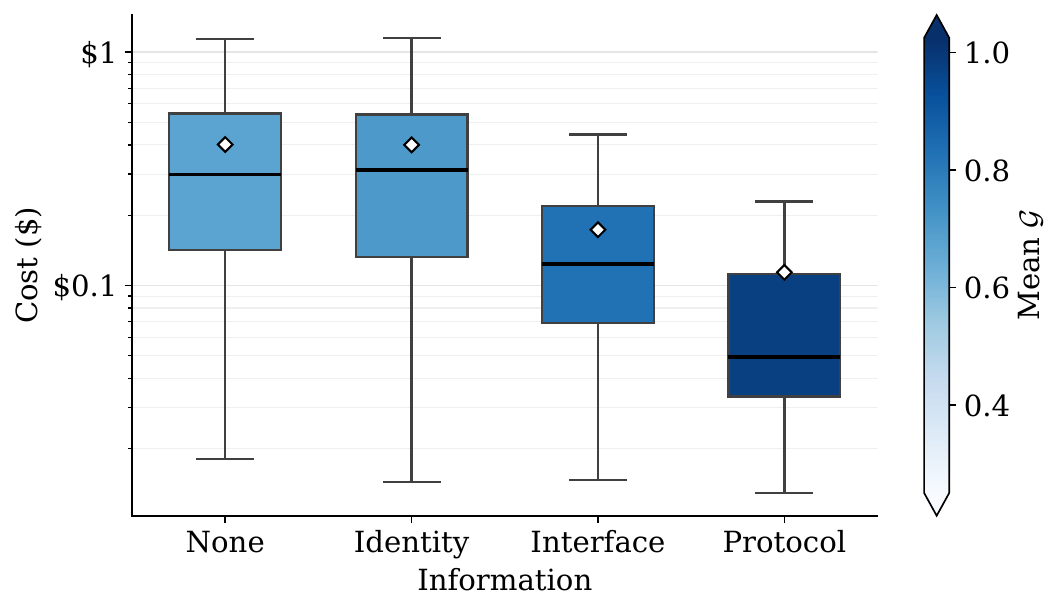}
      \caption{Axis effect of \emph{Information} on task cost.}
      \label{fig:information-cost_axis-effect}
  \end{subfigure}
  \caption{\textit{Left}: $\mathcal{G}$ and number of completed runs per \emph{Model} and \emph{Budget}; error bars show 95\% CI of the mean. \textit{Right}: distribution of a run's cost per \emph{Information} level, colored by that level's mean $\mathcal{G}$. Additional plots are in \cref{fig:information-budget-time-appendix}.}
  \label{fig:model-budget-cost}
\end{figure}

\paragraph{The effect of one axis depends on the others.}
Increasing the \emph{Budget} for the small and medium \emph{Model} raises the completion rate but lowers $\mathcal{G}$, while for the large \emph{Model} it raises both (\cref{fig:model-budget_interactions}). \emph{Information} shows a similar pattern: at lower levels, more \emph{Budget} raises the completion rate without raising $\mathcal{G}$, while at the \emph{interface} level it raises both (\cref{fig:information-budget_interaction}). These patterns largely hold when we compare only configurations that complete under every budget, and the largest interaction overall is between \emph{Information} and \emph{Model} (\cref{sec:apx-completion-robust,sec:apx-effect-sizes}).
One explanation is that additional \emph{Budget} only improves the score once the agent can use it productively, either because it knows what to do (higher \emph{Information}) or because it is capable enough (a larger \emph{Model}). Without either, more time may mainly let the agent continue an unproductive approach until it submits, which would raise the completion rate without raising the score. The \texttt{redshift-estimation} case study matches this explanation: at the \emph{none} level, more \emph{Budget} lowers the score (\cref{tab:budget-info}). One axis can therefore change not only how strongly another axis matters, but also the direction of its effect, and whether it moves completion and score in the same way. This supports evaluating agents as configurable systems rather than fixed models: how the parts combine matters as much as any single part.

\paragraph{Axes affect more than completion and score.}
Agents should be evaluated not only on their score but also on accompanying metrics such as cost and runtime \citep{kapoor2025ai}. Lower \emph{Information} levels not only reach a lower $\mathcal{G}$, but also cost more and run longer (\cref{fig:information-cost_axis-effect,fig:information-time_axis-effect}). We also measure calibration error: the score the agent predicts for itself minus its actual score, normalized by $|\mathcal{R}-\mathcal{B}|$, so that positive values mean the agent overestimates its result.\footnote{Agents report a predicted score only at the \emph{reported} and \emph{binding} levels of \emph{Verification}.} On average, agents are overconfident. \emph{Reasoning} has virtually no effect on calibration error, and the small model is more overconfident than the medium and large models, which behave similarly. More \emph{Information} reduces the calibration error, most strongly at the \emph{protocol} level, where both its mean and its spread drop sharply (\cref{fig:model-information-reasoning-calibration_axis-effect,sec:apx-calibration-error}). \emph{Information} therefore improves $\mathcal{G}$, lowers cost and runtime, and improves calibration at the same time, with no trade-off, making it worth the effort to provide.

\paragraph{Agents rarely decline a weaker specialist.}
On the \textsc{gap\_negative} task \texttt{mmlu-astronomy}, agents should answer with their backbone instead of the weaker specialist. Yet 99.8\% of runs use AstroSage, including almost all runs at the \emph{none} level, where it is not mentioned, and the mean accuracy stays close to the specialist's (\cref{sec:apx-mmlu}). 
Even with the oracle tool (\cref{sec:oracle-ablation}), which lets the agent check the accuracy of its answers, performance does not increase substantially (\cref{tab:oracle-ablation}). Possible explanations are that agents assume a specialist model outperforms a general one, similar to an authority bias~\citep{xie2024adaptive}, or that they explore too little, e.g., by never testing the backbone's own answers. Investigating these is an interesting direction for future work.

\paragraph{Summary.}
Agent outcomes on long, specialized tasks are noisy: repeated runs of the same configuration often disagree on whether the agent finishes, and about five runs per configuration are needed to compare their performance reliably. Completion and outcome consistency also vary sharply between tasks. 
Judging an axis only by its effect on $\mathcal{G}$ is also incomplete. \emph{Information} has the largest effect on $\mathcal{G}$, and at the same time lowers cost and runtime and improves calibration. \emph{Budget}, in contrast, mainly raises $\mathcal{G}$ when the agent has enough information or a large enough model to use the extra time productively. Agents also rarely question a weaker specialist. Building and evaluating agents for such tasks should therefore track completion, score, cost, runtime, and calibration jointly across the axes that shape them, rather than optimizing a single axis for a single metric.

\subsection{Ablations}
\label{sec:ablations}

\paragraph{Different model families behave similarly, but are not interchangeable.}
We rerun the experimental grid with an open-weight model from a different family, Step-3.7-Flash (comparable to the medium Qwen model, Qwen-122B), and with a closed model, Claude Sonnet~5 (roughly comparable to the large Qwen model, Qwen-397B). Both open-weight models rank \emph{Information} first, while the order of the other axes shifts slightly (\cref{tab:model-family-axis-effects}). Step-3.7-Flash also reaches a higher $\mathcal{G}$ than Qwen-122B at about half the cost per run (\cref{tab:model-family-cost}). The comparison between open-weight and closed model shows a sharper difference. The order of the other axes stays the same, but \emph{Budget} rises from third to first rank for Sonnet~5 (\cref{tab:model-family-l-axis-effects}). \emph{Information} still matters, but less than time: the closed model is markedly more sensitive to how long it runs than to how much it is told. Sonnet~5 also executes far more efficiently: pooled across tasks, it needs less than half the tool calls of Qwen-397B, produces about a fifth of the tool-call errors, and is better calibrated. Yet at list prices, it costs more, and its pooled $\mathcal{G}$ is lower (\cref{tab:model-family-l-cost}). Efficient, low-error execution thus does not by itself lead to a better outcome, analogous to the decoupling of capability and reliability described by \citet{rabanser2026towards}. A system tuned to one model's characteristics therefore does not automatically transfer to another, necessitating a systematic analysis like ours to inform the system's priorities.

\paragraph{Verification works better at the system level.}
\label{sec:oracle-ablation}
The \emph{Verification} axis only prompts the agent to check its own work, as a user would, and its effect ranks last among all five axes (\cref{tab:axis-effects}). To test whether verification works better when it is provided by the system instead, we equip the agent with an \emph{oracle} tool that it can call at any point to score its current submission against the task's reference solution. The oracle represents the strongest form of system-level verification; in practice, a system would offer weaker checks, such as a validation set. We rerun the full grid with the oracle for two models, Qwen-122B and Step-3.7-Flash, with all axes varied as in the main experiments (\cref{tab:oracle-ablation}). The oracle raises the score for both models on every task. It also sharply reduces the calibration error, which the prompted \emph{Verification} axis does not. These gains are not free: runtime and cost rise in every case, because the agent makes more tool calls. Since the agent could also use the oracle to search for the correct answer, we checked all oracle runs for such hill-climbing. It occurs in only 4.4\% of runs, and excluding them barely changes the results (\cref{sec:apx-oracle-usage}). This indicates that some agent behaviors, and the outcomes that depend on them, are easier to influence through the system than through the user's prompt.

\paragraph{The findings transfer to a third domain.}
\label{sec:medical}
To test whether our findings depend on the two domains of the main experiments, we further investigate two tasks from medicine with the same structure (\cref{sec:apx-medical}): in \texttt{ecg-diagnosis}, the specialist ECGFounder \citep{li2025electrocardiogram} is stronger than the backbone, and in \texttt{medqa}, the specialist BioMistral-7B \citep{labrak2024biomistral} is weaker (as in \texttt{mmlu-astronomy}). We vary \emph{Information}, \emph{Budget}, and \emph{Model} with five runs per cell and compare against the same subset of the four main tasks. Our findings reappear (\cref{tab:medical-slice}): \emph{Information} has the largest effect on both tasks, and repeated runs disagree on whether the agent completes. How unreliable a single run is depends on the specific task, but repeated runs remain necessary. Agents again trust a weaker specialist although they should know better: on \texttt{medqa}, all runs use the weaker BioMistral and adopt its answers. 
Finally, the interactions between axes reappear: with more budget, $\mathcal{G}$ again falls for the small model and rises for the large one. 

\section{Analysis: Taxonomy of Agent Behavior}
\label{sec:outcome-taxonomy}

Scores show whether an agent succeeds, but not how it behaves or why it fails. To study this, we automatically extract and classify the agents' outcomes and the behaviors that lead to them from the trajectories, using the following pipeline:

\paragraph{Stage 1: Structured Per-Trajectory Extraction.} For every trajectory, we prompt a judge model, DeepSeek-V4-Flash-0731, to describe the trajectory along six dimensions: (i) an overall \textit{result} characterization, (ii) the single root \textit{error category} when applicable, (iii) specialist \textit{model usage}, (iv) \textit{verification behavior}, (v) \textit{planning and exploration}, and (vi) \textit{execution quality}. For every dimension, the judge receives the same context: the task description, the complete trajectory log, automatically derived facts about model usage and scoring, the task's intended solution approach, and the agent's final workspace. 
Following the extract-then-cluster method of \citet{chirkova2026llm}, each dimension is extracted as domain-agnostic free text rather than assigned from a predefined set of labels. The prompts are in \cref{sec:apx-taxonomy-prompts}.

\paragraph{Stage 2: Clustering and Category Induction.} We then consolidate the free-text descriptions of each dimension into a small set of general behavioral categories in three steps. First, we embed the descriptions with a Sentence-BERT model and cluster them with HDBSCAN; points that HDBSCAN leaves unassigned are added to their most similar cluster. Second, the judge model labels each cluster with a short, general description of the behavior, based on a representative sample of its members. Task-specific details go into a separate criterion rather than into the label. Third, the judge model reviews all cluster labels of one dimension at once and merges them into at most $K=8$ general categories, combining near-duplicates and overly specific labels. In practice, no dimension reached this limit. As a result, every trajectory receives one category per dimension, which we cross-tabulate against the experimental conditions (details in \cref{sec:apx-taxonomy}).

\paragraph{Results.}
\begin{figure}[t]
    \centering
    \includegraphics[width=\textwidth]{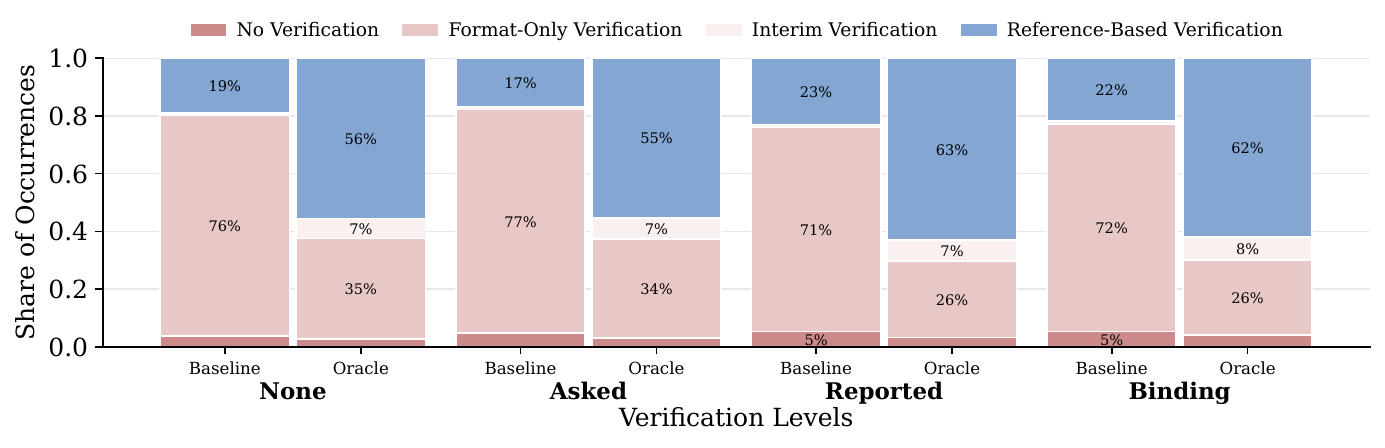}
    \caption{Distribution of verification behavior categories per level of the \emph{Verification} axis, without (Baseline) and with the \emph{oracle} tool, pooled across Qwen-122B, Step-3.7-Flash, and all tasks.}
    \label{fig:verification-behaviour-verification}
\end{figure}

Across all dimensions, \emph{Information} has the strongest association with the distribution of categories (\cref{tab:taxonomy-association}), consistent with its dominant effect on scores. We highlight one case study that connects to the previous section: \emph{verification behavior}. The prompted \emph{Verification} axis had the smallest effect on the score (\cref{tab:axis-effects}), while the oracle tool raised the score for both models on every task (\cref{sec:oracle-ablation}). The taxonomy shows how the agents' behavior differs between the two. Without the oracle, prompting barely changes how agents verify: from \emph{none} to \emph{binding}, the share of reference-based verification rises only from 19\% to 22\%, and about three quarters of runs check only the format of their answer (\cref{fig:verification-behaviour-verification}). With the oracle, the share of reference-based verification roughly triples at every prompt level. Part of this shift reflects the oracle calls themselves, but it shows that agents use a verification option when the system offers one, largely regardless of what the prompt asks for. This illustrates a broader point: some agent behaviors are hard to shift through the user's prompt but respond strongly to what the system provides, and the system should be designed with this distinction in mind.

\section{Conclusion}
\label{sec:discussion}

We systematically study how five axes of an agent's configuration influence its outcomes on scientific tasks that require operating specialist models. Four findings stand out. 
First, a single run is not a reliable measure of a configuration: repeated runs often disagree on whether the agent finishes. Even among genuine attempts about half of the score variance comes from repeating the same configuration, and about five runs per configuration are needed to compare performance reliably.
Second, in our main experiments, the information given to the agent has the largest effect on the score, outranking both the backbone model and the budget, and it also lowers cost and runtime and improves calibration. Third, axes interact: additional budget mainly raises the score when the agent has enough information or a large enough model to use it productively, and can lower the score otherwise, so axes cannot be tuned in isolation. Fourth, some agent behaviors, such as self-verification, barely respond to prompting by the user but shift strongly when the system provides a dedicated tool. Together, these results argue for evaluating and building agents as configurable systems rather than fixed models: tracking completion, variance, cost, and calibration alongside raw scores, and designing the system around behaviors that prompting alone cannot reliably control.

\paragraph{Limitations.}
Our analysis covers over 18{,}000 agent trajectories, but the full grid covers only four tasks across two scientific domains. While each task is designed as a complete factorial experiment and our central findings hold on every task, additional tasks can add further instances of these effects and would strengthen the generality of the conclusions on how far specific rankings generalize. 
We exclude several further system components (memory, multi-agent coordination) that future work should vary as additional axes. Finally, our finding that some agent behaviors respond more to system-level design than to user prompting is demonstrated on one dimension (verification) and needs to be tested across other behavioral dimensions before it can be treated as a general principle.

\subsection*{AI use statement}


In this work, we used generative AI tools to propose or refine hypotheses, design or provide feedback on research methodology or experiments, implement methods, support qualitative and thematic data analysis, and interpret results.
We have not used generative AI tools (or it is not applicable to this work) to generate synthetic data sets, help develop theoretical models or conceptual frameworks, formulate mathematical claims, provide critical ingredients for proving mathematical claims or assist in the writing of proofs.
Additionally, we used generative AI tools to create or modify scientific figures or images, suggest experimental parameters, create or edit software code, draft parts of a research paper, summarize or analyse existing literature, discover research topics or identify gaps, brainstorming, sourcing/searching for information, identify relevant literature, and propose a title or keywords for a research paper. 
We have reviewed all AI-assisted work. All AI use was exclusively used as a first step, and went through manual checks for accuracy, correctness, potential plagiarism, and understandability. We take responsibility for the final content of this work,
including text, claims or artifacts produced with the aid of generative AI.





\subsection*{Reproducibility statement}


To make the paper fully reproducible, we will release all code and instructions on how to use it. We will additionally release all produced artifacts (agent traces) and derived analyses (raw taxonomy data) to encourage further usage in the community.



\subsubsection*{Acknowledgments}
This work was partially funded by the ERC (853489 - DEXIM) and the Alfried Krupp von Bohlen und Halbach Foundation, which we thank for their generous support.
Cordelia Schmid would like to acknowledge the support by the K\"orber European Science Prize and the TUM-IAS Hans Fischer Senior Fellowship.
The authors gratefully acknowledge the Gauss Centre for Supercomputing e.V. (www.gausscentre.eu) for funding this project by providing computing time on the Supercomputer JUPITER at Jülich Supercomputing Centre (JSC). 
We also acknowledge the use of the HPC cluster at Helmholtz Munich for the computational resources used in this study.

\bibliography{iclr2027_conference}
\bibliographystyle{iclr2027_conference}

\appendix
\newpage
\section{Appendix}
\label{sec:appendix}

\subsection{Implementation Details}
We serve each open-weight model with vLLM, using tensor parallelism across its GPUs: one NVIDIA H100 GPU for the small Qwen model, two for the medium, eight for the large, and four for Step-3.7-Flash. All models are FP8-quantized and use a context window of 262{,}144 (Qwen) or 256{,}000 (Step) tokens. Claude Sonnet~5 is accessed through the Anthropic API. Each cell runs as a separate SLURM array job against the shared model server, with up to 8 concurrent agent runs, each in its own sandbox with access to a single GPU. \cref{tab:trajectory-counts} lists the composition of the released trajectories.

\begin{table}[h]
    \centering
    \caption{Composition of the released trajectories. Each cell is repeated 5 times and the grid covers all four tasks.}
    \label{tab:trajectory-counts}
    \small
    \begin{tabular}{@{}lrrr@{}}
        \toprule
        Grid & Cells per task & Runs per task & Trajectories \\
        \midrule
        Main experiments (3 Qwen3.5 models) & 432 & 2{,}160 & 8{,}640 \\
        Step-3.7-Flash (no \emph{act-only}) & 96 & 480 & 1{,}920 \\
        Claude Sonnet~5 & 144 & 720 & 2{,}880 \\
        Oracle, Qwen-122B & 144 & 720 & 2{,}880 \\
        Oracle, Step-3.7-Flash & 96 & 480 & 1{,}920 \\
        \midrule
        Total & & & 18{,}240 \\
        \bottomrule
    \end{tabular}
\end{table}

\subsection{Task Selection}
\label{sec:apx-task-selection}
We selected the tasks through a systematic literature search. For multiple scientific domains (including astronomy, genomics, materials science, physics, medicine, and climate science), we first queried Asta (\url{https://asta.allen.ai}, e.g., ``Foundation models for AI in physics'') for an initial set of seed papers. Among these, we identified papers that release a domain-specific foundation model, as well as survey and review papers. From the survey and review papers, we extracted further references that describe domain-specific foundation models and retrieved these papers as well (backward snowballing). We then annotated all 550 identified papers automatically and used these annotations to classify each paper as a candidate or not, which produced 255 candidate papers. For each candidate, we checked whether the model's weights and code are available and how well it performs. From the remaining models, we chose four tasks that each stress a different step of using a specialist model (pre-processing, fine-tuning, post-processing, and declining a weaker specialist), covering two domains with two tasks each. The selection therefore aims to cover the different demands of operating a specialist model, not science as a whole.

\subsection{Tool Use and Specialist Model Usage}
\label{sec:apx-tool-use}
\cref{tab:tool-calls} shows, for each tool, the mean number of calls per run and the share of these calls that returned an error, for the main experiments. A \texttt{run\_bash} call counts as an error if the command does not exit with code 0. A call to any other tool counts as an error if the harness reports that it could not execute it (e.g., a missing file or an unreachable URL). \emph{Other/malformed} covers all calls to tool names the harness did not recognize, which always fail. \texttt{run\_bash} accounts for most calls on every task, while the use of \texttt{web\_fetch} differs substantially between tasks.
\begin{table}[h]
    \centering
    \caption{Mean tool calls per run and error rate per tool, main experiments.}
    \label{tab:tool-calls}
    \small
    \resizebox{\textwidth}{!}{%
    \begin{tabular}{@{}lrrrrrrrrrrrrrr@{}}
        \toprule
        & \multicolumn{2}{c}{\texttt{run\_bash}} & \multicolumn{2}{c}{\texttt{read\_file}} & \multicolumn{2}{c}{\texttt{write\_file}} & \multicolumn{2}{c}{\texttt{web\_fetch}} & \multicolumn{2}{c}{\texttt{finish}} & \multicolumn{2}{c}{Other/malformed} & \multicolumn{2}{c}{Total} \\
        \cmidrule(lr){2-3}\cmidrule(lr){4-5}\cmidrule(lr){6-7}\cmidrule(lr){8-9}\cmidrule(lr){10-11}\cmidrule(lr){12-13}\cmidrule(lr){14-15}
        Task & Calls & Err.\% & Calls & Err.\% & Calls & Err.\% & Calls & Err.\% & Calls & Err.\% & Calls & Err.\% & Calls & Err.\% \\
        \midrule
        \texttt{redshift-estimation} & 28.51 & 11.4 & 4.24 & 5.7 & 3.56 & 3.9 & 0.79 & 0.0 & 0.76 & 0.1 & 0.07 & 100.0 & 37.93 & 9.7 \\
        \texttt{mmlu-astronomy} & 17.84 & 12.7 & 2.92 & 1.3 & 2.36 & 1.4 & 0.07 & 3.8 & 0.95 & 0.0 & 0.05 & 100.0 & 24.20 & 9.9 \\
        \texttt{promoter-prediction} & 16.31 & 16.2 & 3.09 & 0.7 & 3.73 & 3.7 & 0.04 & 4.3 & 0.89 & 0.1 & 0.04 & 100.0 & 24.10 & 11.8 \\
        \texttt{rna-folding} & 33.21 & 14.2 & 3.73 & 4.0 & 6.14 & 3.4 & 1.62 & 0.7 & 0.62 & 0.3 & 0.18 & 100.0 & 45.49 & 11.6 \\
        \bottomrule
    \end{tabular}%
    }
\end{table}

\cref{tab:web-fetch-hosts} breaks the \texttt{web\_fetch} calls down by destination. \emph{Ref.\ correct} and \emph{Ref.\ decoy} give the share of fetches to a URL listed in the README of the correct specialist model or of a decoy. Decoys are the other specialist models in the sandbox, which do not fit the task; on \texttt{mmlu-astronomy}, we also count AstroSage as a decoy, because the correct behavior is not to use it. \emph{Off-list} covers all other URLs. On the three \textsc{gap\_positive} tasks, 85--96\% of fetches go to the correct model's reference material. The off-list remainder is mostly general troubleshooting (a search engine, Colab, PyPI, the GitHub API), except on \texttt{promoter-prediction}, where some off-list fetches to \texttt{huggingface.co} read similarly named but wrong repositories (e.g., \texttt{nlpaueb/dnabert-2-base-100M}) instead of the one provided. On \texttt{mmlu-astronomy}, most fetches (73\%) also go to AstroSage's reference material, while 27\% go elsewhere, mainly to a search engine and Wikipedia.
\begin{table}[h]
    \centering
    \caption{\texttt{web\_fetch} destinations, main experiments. HTTP OK \% is the share of fetches that return a 2xx status. A fetch that returns 404 still counts as successful in \cref{tab:tool-calls}, because the harness executed it without an error.}
    \label{tab:web-fetch-hosts}
    \small
    \resizebox{\textwidth}{!}{%
    \begin{tabular}{@{}lrrrrrl@{}}
        \toprule
        Task & Fetches/run & Ref.\ correct \% & Ref.\ decoy \% & Off-list \% & HTTP OK \% & Top off-list hosts \\
        \midrule
        \texttt{redshift-estimation} & 0.79 & 88.9 & 0.1 & 11.0 & 84.4 & \texttt{colab.research.google.com}, \texttt{lite.duckduckgo.com} \\
        \texttt{mmlu-astronomy} & 0.07 & 0.0 & 73.0 & 27.0 & 87.4 & \texttt{lite.duckduckgo.com}, \texttt{en.wikipedia.org} \\
        \texttt{promoter-prediction} & 0.04 & 84.8 & 0.0 & 15.2 & 85.9 & \texttt{huggingface.co}, \texttt{raw.githubusercontent.com} \\
        \texttt{rna-folding} & 1.62 & 95.9 & 0.0 & 4.1 & 86.7 & \texttt{api.github.com}, \texttt{lite.duckduckgo.com} \\
        \bottomrule
    \end{tabular}%
    }
\end{table}

\cref{tab:model-usage} shows how agents engage with each specialist model in the main experiments, based on patterns in the trajectories. It gives the share of runs in which a model's directory is \emph{mentioned} in a tool call, its documentation is read (\emph{read\_docs}), a command that names the model's path runs and exits with code 0 (\emph{executed}), and the agent fetches a URL from the model's README (\emph{researched\_online}). 
In the main experiments, the sandbox contains every specialist model that was part of the framework when the task's grid was run: the two astronomy models (AstroCLIP and AstroSage) for mmlu-astronomy and redshift-estimation, and these two plus the two genomics models (DNABERT-2 and RiNALMo) for promoter-prediction and rna-folding.
In all ablations, every task has these four models available, and the two medical tasks additionally have ECGFounder and BioMistral, i.e., six models (\cref{sec:apx-medical}).
On the three \textsc{gap\_positive} tasks, the correct specialist model is read and executed in 95--99\% of runs, while no decoy is executed in more than 1\% of runs. \texttt{mmlu-astronomy} has no correct specialist model, since it can be answered directly by the backbone. Still, the weaker AstroSage is read in 98.6\% of runs and executed in 66.1\%. Because \emph{executed} only counts commands that name the model's path, it undercounts use: counting scripts that load AstroSage as well, 99.8\% of runs use it (\cref{sec:apx-mmlu}).
\begin{table}[h]
    \centering
    \caption{Engagement with each specialist model (\% of runs), main experiments. The correct specialist model per task is in bold, \texttt{mmlu-astronomy} has none. \texttt{redshift-estimation} and \texttt{mmlu-astronomy} were run before the genomics models were available to the agent, however, the genomics tasks, whose sandbox contained all four models, show that the additional models do not keep the agent from finding the right one.}
    \label{tab:model-usage}
    \small
    \begin{tabular}{@{}llrrrr@{}}
        \toprule
        Task & Specialist & \textit{mentioned} & \textit{read\_docs} & \textit{executed} & \textit{researched\_online} \\
        \midrule
        \multirow{4}{*}{\texttt{redshift-estimation}}
          & \texttt{\textbf{astroclip}} & 100.0\% & 99.9\% & 99.3\% & 28.3\% \\
          & \texttt{astrosage} & 27.3\% & 27.1\% & 0.2\% & 0.1\% \\
          & \texttt{dnabert-2} & 0.0\% & 0.0\% & 0.0\% & 0.0\% \\
          & \texttt{rinalmo} & 0.0\% & 0.0\% & 0.0\% & 0.0\% \\
        \addlinespace
        \multirow{4}{*}{\texttt{mmlu-astronomy}}
          & \texttt{astroclip} & 30.5\% & 30.2\% & 0.3\% & 0.0\% \\
          & \texttt{astrosage} & 100.0\% & 98.6\% & 66.1\% & 5.0\% \\
          & \texttt{dnabert-2} & 0.0\% & 0.0\% & 0.0\% & 0.0\% \\
          & \texttt{rinalmo} & 0.0\% & 0.0\% & 0.0\% & 0.0\% \\
        \addlinespace
        \multirow{4}{*}{\texttt{promoter-prediction}}
          & \texttt{\textbf{dnabert-2}} & 100.0\% & 99.8\% & 94.7\% & 3.6\% \\
          & \texttt{astroclip} & 11.9\% & 11.9\% & 0.0\% & 0.0\% \\
          & \texttt{astrosage} & 10.1\% & 10.1\% & 0.0\% & 0.0\% \\
          & \texttt{rinalmo} & 24.3\% & 24.3\% & 0.2\% & 0.0\% \\
        \addlinespace
        \multirow{4}{*}{\texttt{rna-folding}}
          & \texttt{\textbf{rinalmo}} & 100.0\% & 100.0\% & 99.3\% & 38.2\% \\
          & \texttt{astroclip} & 10.3\% & 9.3\% & 0.1\% & 0.0\% \\
          & \texttt{astrosage} & 8.9\% & 8.3\% & 0.0\% & 0.0\% \\
          & \texttt{dnabert-2} & 14.6\% & 14.0\% & 0.1\% & 0.0\% \\
        \bottomrule
    \end{tabular}
\end{table}

\subsection{Variance Decomposition and Hurdle}
\label{sec:apx-noise-floor}
\paragraph{Intraclass correlation.}
To separate real signal between configurations from run-to-run noise, we decompose the score variance in two steps. First, a one-way random-effects ANOVA over all cells with at least three completed runs splits the total variance into a within-cell component ($MS_{\text{within}}$) and a bias-corrected between-cell component. From these, we compute the intraclass correlation
\[
\text{ICC} = \frac{\sigma^2_{\text{between}}}{\sigma^2_{\text{between}} + \sigma^2_{\text{within}}},
\]
the share of the total variance that comes from differences between cells rather than from run-to-run noise. On its own, this decomposition is misleading: a run can complete and still fail to make a genuine attempt at the task (e.g., by running the wrong specialist model or submitting predictions far below the trivial score). Because the ANOVA sums squared deviations, such rare but severe failures inflate $MS_{\text{within}}$ far out of proportion to how often they occur. We therefore also compute a two-part (hurdle) decomposition~\citep{cragg1971}, following recent work on the reliability of LLM agents across repeated trials that separates a binary success indicator per run from a continuous score given success~\citep{rabanser2026towards,khanal2026beyond}.

\begin{table}[h]
  \centering
  \caption{Hurdle decomposition and run-to-run variance per task. Mean per-cell SD and the 95\% CI half-width of a cell's mean score (at 5 runs) are expressed as \% of $|\mathcal{R}-\mathcal{T}|$, so they are comparable across tasks. The pooled CI half-width in units of $\mathcal{G}$ is shown in \cref{fig:variance-appendix} (right).}
  \label{tab:hurdle-decomposition}
  \small
  \resizebox{\textwidth}{!}{%
  \begin{tabular}{@{}lrrrrrr@{}}
      \toprule
      Task & Hurdle failure rate (95\% CI) & $C_{\text{out}}$ & ICC, all runs & ICC, clears hurdle & Mean per-cell SD & 95\% CI (5 runs) \\
      \midrule
      \texttt{redshift-estimation} & 53.8\% [51.4, 56.2] & 0.727 & 33.3\% & 36.2\% & 85.1\% & 91.9\% \\
      \texttt{mmlu-astronomy}      & 4.7\%  [3.9, 5.7]   & 0.905 & 29.4\% & 6.9\%  & 17.6\% & 20.8\% \\
      \texttt{promoter-prediction} & 0.5\%  [0.2, 0.9]   & 0.986 & 47.9\% & 49.7\% & 10.7\% & 12.9\% \\
      \texttt{rna-folding}         & 8.7\%  [7.3, 10.3]  & 0.825 & 45.7\% & 35.2\% & 12.8\% & 13.0\% \\
      \midrule
      Pooled (3 \textsc{gap\_positive} tasks, task means removed) & 20.6\% [19.5, 21.7] & 0.858 & 39.4\% & 46.1\% & --- & --- \\
      \bottomrule
  \end{tabular}%
  }
\end{table}

\paragraph{Hurdle indicator and failure rate.} We write $y \in \{0,1\}$ for the hurdle indicator $\mathcal{H}$ of a run: $y=1$ if the run's score exceeds the trivial score $\mathcal{T}$ by more than the margin $\mu$, and $y=0$ otherwise. The hurdle is distinct from completion: runs that do not complete (e.g., because they run out of budget) have no score and are not counted here at all. The \emph{hurdle failure rate} is the share of completed runs with $y=0$, i.e., $1-\bar{y}$. We report 95\% Wilson score intervals~\citep{wilson1927}. 
The completion rate in \cref{tab:results-overview} uses the same interval, computed over all 2{,}160 runs of a task. The $\pm$ values in \cref{tab:results-overview} give the half-widths of these intervals, which are nearly symmetric at these sample sizes.

\paragraph{Outcome consistency.} For each cell $i$ with at least 3 completed runs, let $\hat p_i$ be the share of its runs that clear the hurdle. Following~\citet{rabanser2026towards},
\[
C_{\text{out}} = \frac{1}{k}\sum_{i=1}^{k} (2\hat p_i - 1)^2 = \frac{1}{k}\sum_{i=1}^{k}\left[1 - \frac{\hat p_i(1-\hat p_i)}{0.25}\right],
\]
averaged over the $k$ qualifying cells. Each term normalizes the variance of the cell's binary outcomes by the maximum variance of a Bernoulli variable, 0.25. A cell contributes 1 if all its runs clear the hurdle or all fail it, and 0 if exactly half of them clear it. $C_{\text{out}} = 1$ therefore means that every configuration reliably succeeds or reliably fails.
  
\paragraph{Variance among genuine attempts.} We apply the same ANOVA and ICC to runs with $y=1$ only. This measures how repeatable an agent's score is, given that it makes a genuine attempt. Since this restriction selects runs by their outcome, it can bias the variance estimate, so we report both versions.

\paragraph{Pooling across tasks.} Cells are defined per task. Pooling the raw $\mathcal{G}$ values of several tasks would therefore count differences between the task means as differences between cells, and inflate the pooled ICC above every per-task value. We therefore subtract each task's mean before pooling.

\paragraph{Reliability and number of runs.}
The ICC is the reliability of a single run: the share of the variance of single-run scores that reflects real differences between cells \citep{shrout1979intraclass}. The mean of $n$ runs per cell has the reliability
\[
\rho_n = \frac{\sigma^2_{\text{between}}}{\sigma^2_{\text{between}} + \sigma^2_{\text{within}}/n} = \frac{n \cdot \text{ICC}}{1 + (n-1)\,\text{ICC}},
\]
which is the Spearman--Brown formula and the simplest case of a decision study in generalizability theory \citep{brennan1992generalizability}. Solving $\rho_n = 0.8$ for $n$ gives the number of runs per cell that are needed to reach the common reliability threshold of 0.8. \Cref{tab:reliability} reports both quantities. The reliability refers to comparing configurations within one task. \texttt{mmlu-astronomy} needs far more runs among genuine attempts, because almost all of its runs use the same weaker specialist, so the configurations barely differ in score (\cref{sec:apx-mmlu}).

\begin{table}[h]
  \centering
  \caption{Reliability of a cell's mean score with 1 and 5 runs, and the number of runs per cell needed for a reliability of 0.8, computed from the ICCs in \cref{tab:hurdle-decomposition}.}
  \label{tab:reliability}
  \small
  \begin{tabular}{@{}lrrrrrr@{}}
      \toprule
      & \multicolumn{3}{c}{All runs} & \multicolumn{3}{c}{Clears hurdle} \\
      \cmidrule(lr){2-4} \cmidrule(lr){5-7}
      Task & 1 run & 5 runs & Runs for 0.8 & 1 run & 5 runs & Runs for 0.8 \\
      \midrule
      \texttt{redshift-estimation} & 0.33 & 0.71 & 8.0 & 0.36 & 0.74 & 7.1 \\
      \texttt{mmlu-astronomy}      & 0.29 & 0.68 & 9.6 & 0.07 & 0.27 & 54.0 \\
      \texttt{promoter-prediction} & 0.48 & 0.82 & 4.4 & 0.50 & 0.83 & 4.1 \\
      \texttt{rna-folding}         & 0.46 & 0.81 & 4.8 & 0.35 & 0.73 & 7.4 \\
      \midrule
      Pooled (3 \textsc{gap\_positive} tasks) & 0.39 & 0.76 & 6.2 & 0.46 & 0.81 & 4.7 \\
      \bottomrule
  \end{tabular}
\end{table}

\subsection{Additional Completion Analyses}
\label{sec:apx-completion-analysis}
\begin{figure}[h]
    \centering
    \includegraphics[width=0.57\textwidth]{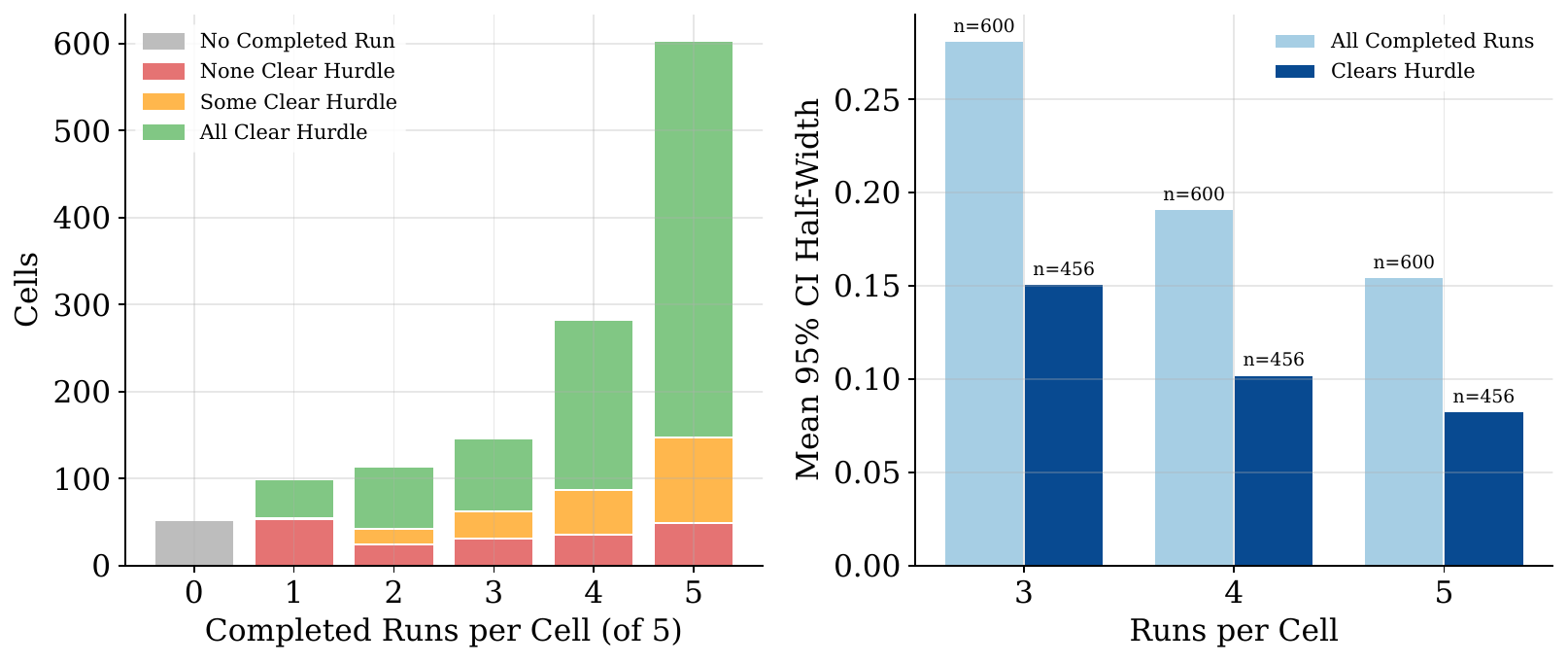}
    \caption{Completion and variance details, pooled across the three \textsc{gap\_positive} tasks. \textit{Left}: cells by their number of completed runs (0--5); color shows whether all, some, or none of a cell's completed runs clear the hurdle (green/orange/red), and gray marks cells without any completed run. \textit{Right}: mean 95\% CI half-width of a cell's mean $\mathcal{G}$ at 3--5 runs, for all completed runs (light blue) and for hurdle-clearing runs only (dark blue).}
    \label{fig:variance-appendix}
\end{figure}

Not every run produces a score, for example when the agent runs out of budget before submitting a solution. Across all four tasks, 80.7\% of runs complete, and 19.3\% run out of budget. Completion also varies within configurations: on every task, a substantial share of cells completes on some runs but not on others, from run-to-run variation alone (\cref{tab:results-overview}; \cref{fig:variance-appendix}, left, for the three \textsc{gap\_positive} tasks). 
Per-cell completion rates range from 0\% to 100\% on every task except \texttt{mmlu-astronomy}, where every configuration completes at least 20\% of its runs.
We flag a run as an extreme outlier if its score differs from the trivial score by more than 10 times the gap between reference and trivial score, $10\,|\mathcal{R}-\mathcal{T}|$. Such outliers occur only on \texttt{redshift-estimation} (7 of 1{,}647 completed runs, 0.4\%), all but one with the small model. We exclude them from all score-based analyses.

\begin{figure}[h]
    \centering
    \begin{subfigure}{0.59\textwidth}
        \centering
        \includegraphics[width=\textwidth]{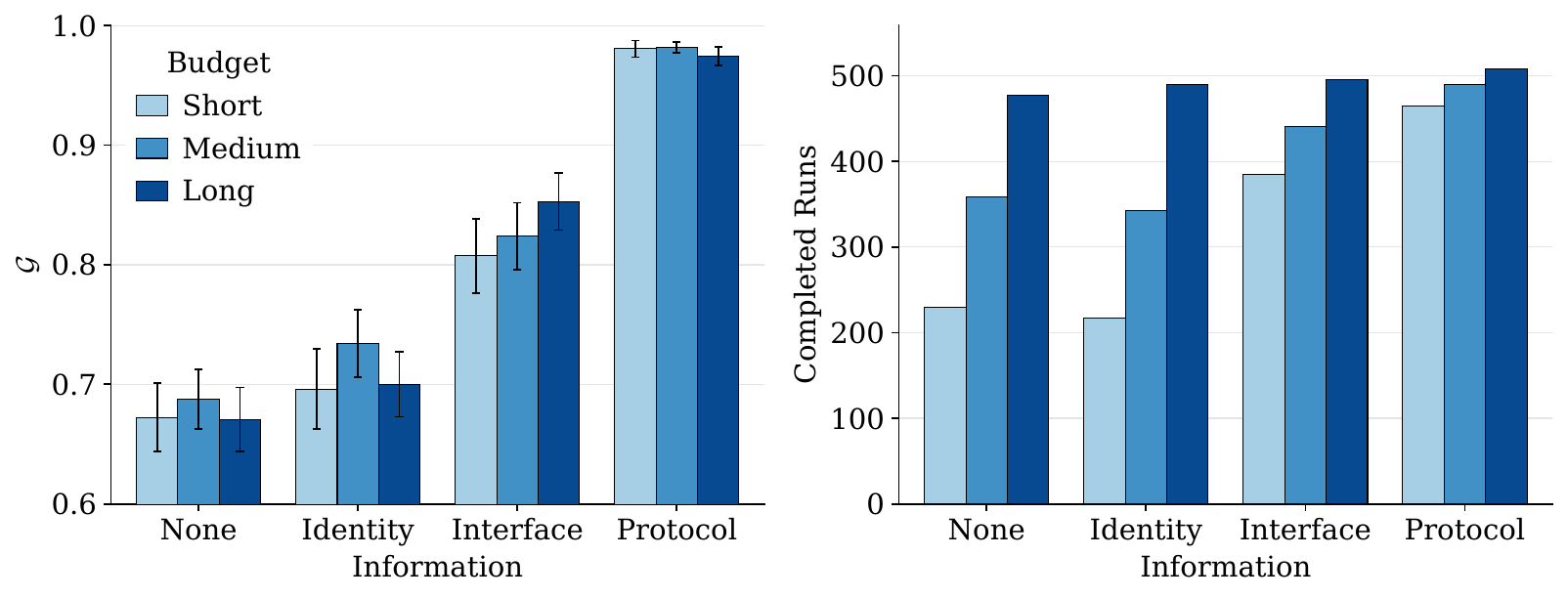}
        \caption{Interaction of the \emph{Information} and \emph{Budget} axes.}
        \label{fig:information-budget_interaction}
    \end{subfigure}
    \hfill
    \begin{subfigure}{0.39\textwidth}
        \centering
        \includegraphics[width=\textwidth]{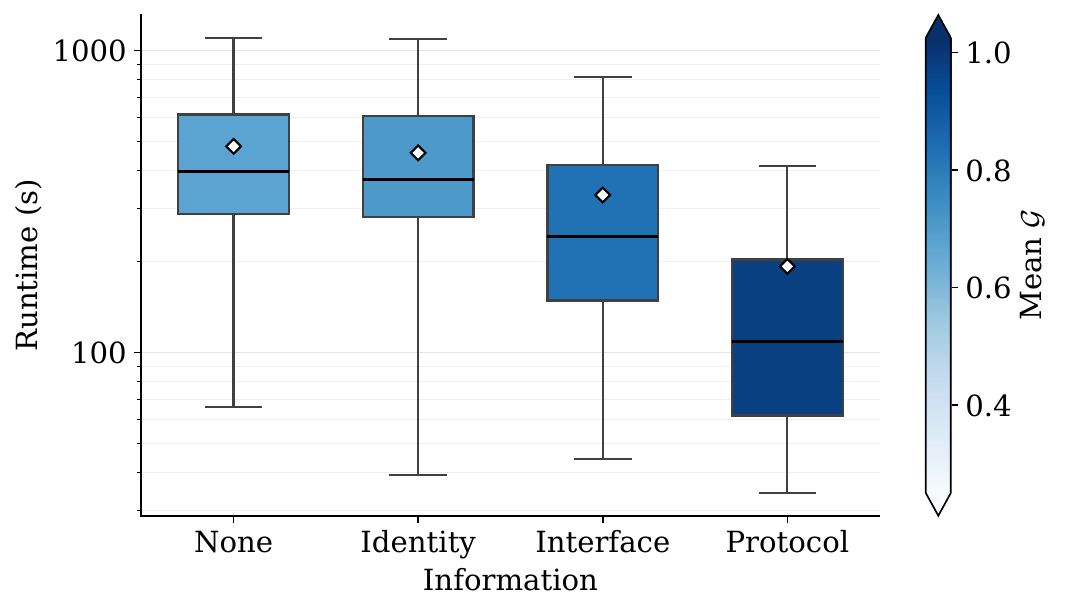}
        \caption{Axis effect of \emph{Information} on runtime.}
        \label{fig:information-time_axis-effect}
    \end{subfigure}
    \caption{\textit{Left}: $\mathcal{G}$ and number of completed runs per \emph{Information} and \emph{Budget} level; error bars show 95\% CI of the mean. \textit{Right}: distribution of a run's wall-clock time per \emph{Information} level, colored by that level's mean $\mathcal{G}$.}
    \label{fig:information-budget-time-appendix}
\end{figure}

\subsection{Additional Interaction and Runtime Views}
\label{sec:apx-interaction}
\Cref{fig:information-budget-time-appendix} shows the \emph{Information}$\times$\emph{Budget} interaction, next to the \emph{Model}$\times$\emph{Budget} interaction in the main text (\cref{fig:model-budget-cost}), and the effect of \emph{Information} on runtime, next to its effect on cost.  

\begin{wraptable}{r}{0.45\textwidth}
    \centering
    \caption{Mean $R^2$ \& $\mathcal{H}$ failure rate, Budget $\times$ Information, \texttt{redshift-estimation}.}
    \label{tab:budget-info}
    \small
    \begin{tabular}{@{}lrrr@{}}
        \toprule
        Info/Budget & Short & Medium & Long \\
        \midrule
        \multirow{2}{*}{None}      & $-0.429$     & $-0.441$     & $-0.713$     \\
                                   & \tiny 89.1\% & \tiny 92.0\% & \tiny 92.2\% \\
        \multirow{2}{*}{Identity}  & $-0.514$     & $-0.519$     & $-0.516$     \\
                                   & \tiny 83.8\% & \tiny 84.4\% & \tiny 84.4\% \\
        \multirow{2}{*}{Interface} & $-0.605$     & $-0.560$     & $-0.380$     \\
                                   & \tiny 76.1\% & \tiny 71.9\% & \tiny 65.9\% \\
        \multirow{2}{*}{Protocol}  & $0.753$      & $0.753$      & $0.753$      \\
                                   & \tiny 0.0\%  & \tiny 0.0\%  & \tiny 0.0\%  \\
        \bottomrule
    \end{tabular}
\end{wraptable}
\paragraph{Case study: \texttt{redshift-estimation}.}
How axes interact depends on the task, but \texttt{redshift-estimation} offers a clear example (\cref{tab:budget-info}). At the \emph{none} level, more \emph{Budget} lowers the score ($-0.429\to-0.713$) at similar hurdle failure rates, consistent with the agent pursuing a wrong approach further. At the \emph{interface} level, more \emph{Budget} raises the score ($-0.605\to-0.380$) and lowers the hurdle failure rate: the agent now knows enough to use the extra time. At the \emph{protocol} level, \emph{Budget} makes no difference, because every run already reproduces the reference score. The medians show the same pattern (\emph{none}: $-0.34\to-0.50$; \emph{interface}: $-0.55\to-0.34$), and so does the comparison restricted to configurations that have completed runs under all three budgets (\emph{none}: $-0.44\to-0.77$; \emph{interface}: $-0.61\to-0.38$).

\subsection{Calibration Error}
\label{sec:apx-calibration-error}
\begin{figure}[h]
    \centering
    \begin{subfigure}{0.30\textwidth}
        \centering
        \includegraphics[width=\textwidth]{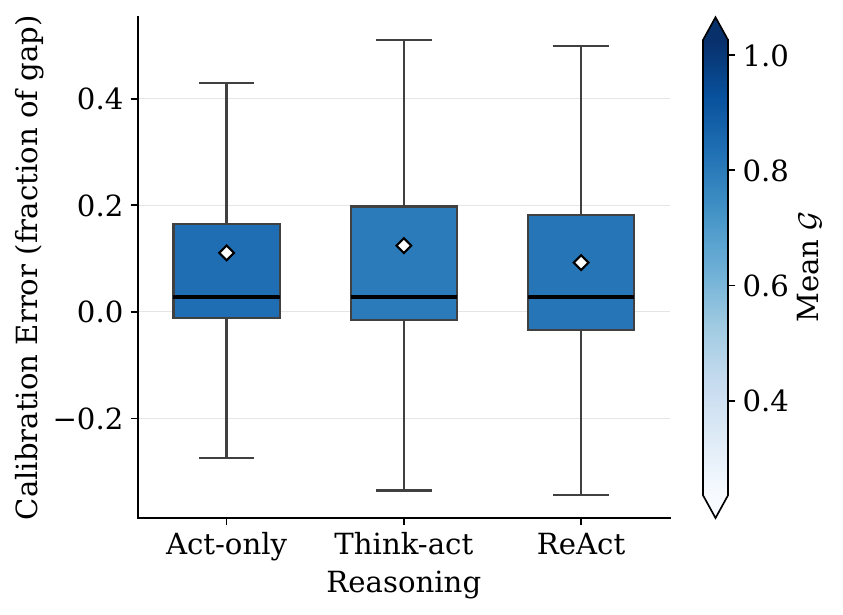}
        \caption{\emph{Reasoning}}
        \label{fig:reasoning-calibration_axis-effect}
    \end{subfigure}
    \hfill
    \begin{subfigure}{0.30\textwidth}
        \centering
        \includegraphics[width=\textwidth]{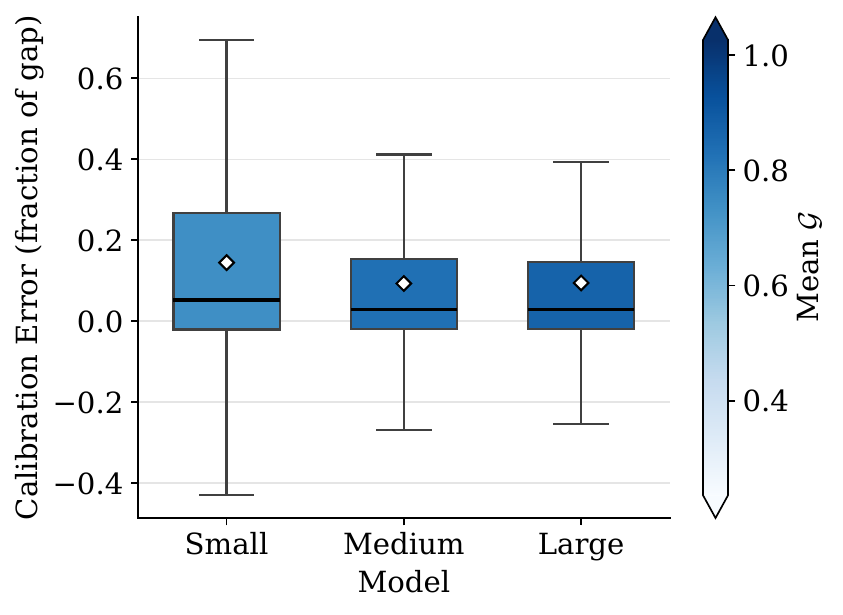}
        \caption{\emph{Model}}
        \label{fig:model-calibration_axis-effect}
    \end{subfigure}
    \hfill
    \begin{subfigure}{0.37\textwidth}
        \centering
        \includegraphics[width=\textwidth]{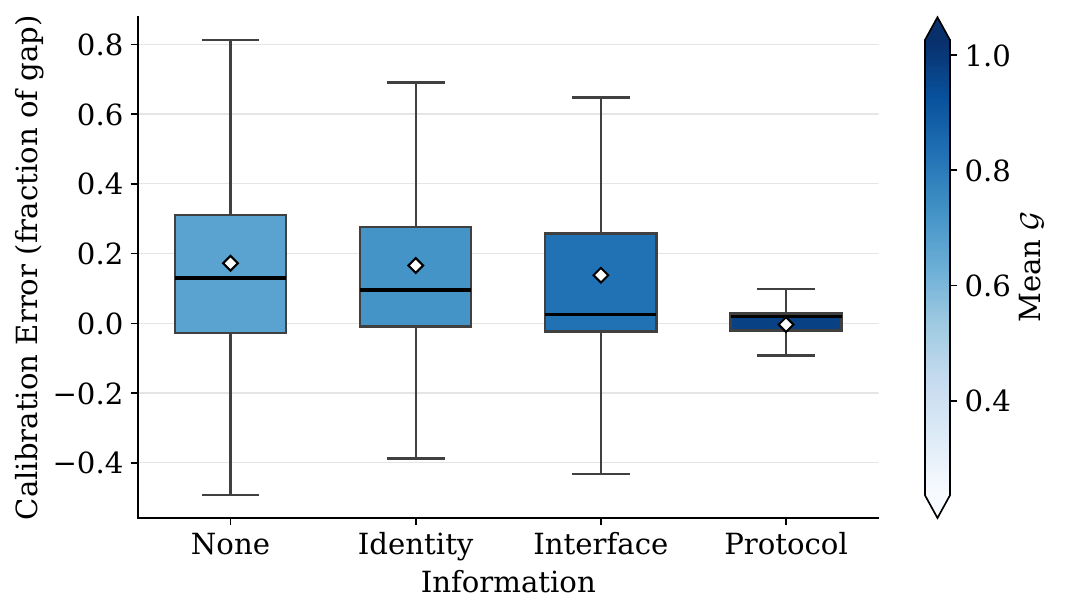}
        \caption{\emph{Information}}
        \label{fig:information-calibration_axis-effect}
    \end{subfigure}
    \caption{Calibration error per level of \emph{Reasoning}, \emph{Model}, and \emph{Information}, pooled across the three \textsc{gap\_positive} tasks. Positive values mean the agent overestimates its score. Each box is colored by that level's mean $\mathcal{G}$.}
    \label{fig:model-information-reasoning-calibration_axis-effect}
\end{figure}

\cref{fig:model-information-reasoning-calibration_axis-effect} shows the calibration error, defined in \cref{sec:results} as the agent's predicted score minus its actual score, normalized by $|\mathcal{R}-\mathcal{B}|$, per level of \emph{Reasoning}, \emph{Model}, and \emph{Information}.

\subsection{Agent Effort and Cost}
\label{sec:apx-cost}
\begin{table}[b]
    \centering
    \caption{Mean cost and effort per run in the main experiments (432 cells $\times$ 5 runs).}
    \label{tab:cost-effort}
    \small
    \resizebox{\textwidth}{!}{%
    \begin{tabular}{@{}lrrrr@{}}
        \toprule
         & \texttt{redshift-} & \texttt{mmlu-} & \texttt{promoter-} & \texttt{rna-} \\
         & \texttt{estimation} & \texttt{astronomy} & \texttt{prediction} & \texttt{folding} \\
        \midrule
        Mean Runtime (s)            & 344.2 & 188.3 & 278.0 & 476.0 \\
        Mean Cost (USD)             & 0.277 & 0.107 & 0.115 & 0.424 \\
        Mean Steps / Tool Calls / Errors           & 33 / 38 / 4   & 21 / 24 / 2    & 20 / 24 / 3   & 41 / 45 / 5    \\
        \bottomrule
    \end{tabular}%
    }
\end{table}
\cref{tab:cost-effort} gives the mean runtime, cost, steps, tool calls, and tool-call errors per run in the main experiments. Since we run the open-weight models on our own GPUs, we estimate their cost from token counts at the first-party API list prices (USD per 1M input/output tokens): \$0.25/\$2.00 for Qwen-35B, \$0.40/\$3.20 for Qwen-122B, and \$0.60/\$3.60 for Qwen-397B (Alibaba Cloud Model Studio), and \$0.20/\$1.15 for Step-3.7-Flash (StepFun). For Claude Sonnet~5, we use Anthropic's list price of \$2/\$10. \cref{tab:anthropic-cost} gives the total cost of one full grid per task and model. For Sonnet~5, \emph{Calculated} prices its tokens at the list price without caching discounts, using the same formula as for the open-weight models, and is therefore the directly comparable number. \emph{Real} is what Anthropic actually billed, with prompt caching, and \emph{Savings} gives how much lower this is than \emph{Calculated}.
\begin{table}[t]
  \centering
  \caption{Cost of one full 720-run task grid, per model. Tokens shown as prompt / completion, in millions.}
  \label{tab:anthropic-cost}
  \footnotesize
  \resizebox{\linewidth}{!}{%
  \begin{tabular}{@{}l lr lr lr lrrr@{}}
      \toprule
      & \multicolumn{2}{c}{Qwen-35B-A3B} & \multicolumn{2}{c}{Qwen-122B-A10B} & \multicolumn{2}{c}{Qwen-397B-A17B} & \multicolumn{4}{c}{Sonnet 5} \\ 
      \cmidrule(lr){2-3} \cmidrule(lr){4-5} \cmidrule(lr){6-7} \cmidrule(lr){8-11} Task & Tokens & Cost & Tokens & Cost & Tokens & Cost & Tokens & Calculated & Real & Savings \\ 
      \midrule
      \texttt{redshift} & 659.5 / 10.3 & \$186 & 397.2 / 7.7 & \$183 & 340.1 / 7.0 & \$229 & 147.8 / 4.1 & \$337 & \$124 & 63.3\% \\ 
      \texttt{mmlu} & 222.3 / 3.7 & \$63 & 142.0 / 3.2 & \$67 & 150.7 / 3.1 & \$102 & 65.4 / 2.8 & \$159 & \$66 & 58.6\% \\ 
      \texttt{promoter} & 306.0 / 7.1 & \$91 & 143.0 / 4.9 & \$73 & 117.4 / 4.1 & \$85 & 88.5 / 3.9 & \$216 & \$84 & 60.9\% \\ 
      \texttt{rna} & 978.5 / 15.3 & \$275 & 579.3 / 13.0 & \$273 & 530.3 / 13.5 & \$367 & 140.8 / 4.6 & \$328 & \$110 & 66.4\% \\ 
      \bottomrule
  \end{tabular}
  }
\end{table}

\subsection{Ablation Details}
\label{sec:apx-ablation-details}
The outcome and cost tables in this subsection (\cref{tab:model-family-cost,tab:model-family-l-cost,tab:oracle-ablation}) share the same conventions. \texttt{mmlu-astronomy} is \textsc{gap\_negative}, so its row reports raw accuracy$^\dagger$ instead of $\mathcal{G}$. 
Calibration error is normalized by $|\mathcal{R}-\mathcal{B}|$. Cost, wall-clock time, steps, tool calls, and errors are means per run, with costs estimated at list prices (\cref{sec:apx-cost}).

\paragraph{Oracle ablation.}
\label{par:apx-oracle-ablation}
This paragraph gives the full results of the oracle ablation (\cref{sec:oracle-ablation}). The agent receives an \texttt{oracle\_check} tool, independent of the prompted \emph{Verification} level, which scores its current submission against the task's reference solution on demand. We rerun the full grid with this tool for two models, Qwen-122B and Step-3.7-Flash, with all axes varied as in the main experiments. \Cref{tab:oracle-ablation} compares outcome, calibration, cost, and effort with and without the tool, and \cref{tab:oracle-axis-effects} compares the axis ranks. The baseline for Qwen-122B uses its full grid, including \emph{act-only}, while \cref{tab:model-family-cost} excludes this level; the numbers for Qwen-122B in the two tables therefore differ slightly.

\paragraph{Model-family and open- vs.\ closed-weight ablations.}
\label{sec:apx-model-family-ablation}
This paragraph gives the full results of the two model ablations in \cref{sec:ablations}. \cref{tab:model-family-axis-effects,tab:model-family-cost} compare the two open-weight models, and \cref{tab:model-family-l-axis-effects,tab:model-family-l-cost} compare an open-weight and a closed model. For each model, we repeat the axis-effect ranking (\emph{Model} is fixed within each block and therefore omitted) and report outcome, calibration, cost, and effort.

\begin{table}[h]
  \centering
  \caption{Oracle ablation: outcome, calibration error, cost/effort with vs.\ without the \texttt{oracle\_check}.}
  \label{tab:oracle-ablation}
  \small
  \resizebox{\textwidth}{!}{%
  \begin{tabular}{@{}llrrrrrrrrrr@{}}
      \toprule
      & & \multicolumn{2}{c}{Outcome} & \multicolumn{2}{c}{Cal. Err.} & \multicolumn{2}{c}{Cost (USD)} & \multicolumn{2}{c}{Wall-clock (s)} & \multicolumn{2}{c}{Steps / TC / Err.} \\
      \cmidrule(lr){3-4} \cmidrule(lr){5-6} \cmidrule(lr){7-8} \cmidrule(lr){9-10} \cmidrule(lr){11-12}
      Model & Task(s) & Baseline & Oracle & Baseline & Oracle & Baseline & Oracle & Baseline & Oracle & Baseline & Oracle \\
      \midrule
      \multirow{2}{*}{Qwen-122B} & \texttt{mmlu-astronomy}$^\dagger$ & 0.707 & 0.725 & 0.152 & $-$0.002 & 0.093 & 0.125 & 165 & 220 & 19 / 22 / 2 & 21 / 24 / 2 \\
       & Pooled \textsc{gap\_positive} & 0.822 & 0.923 & 0.093 & 0.002 & 0.245 & 0.358 & 346 & 453 & 28 / 32 / 4 & 34 / 39 / 5 \\
      \addlinespace
      \multirow{2}{*}{Step-3.7-Flash} & \texttt{mmlu-astronomy}$^\dagger$ & 0.677 & 0.721 & 0.285 & 0.065 & 0.043 & 0.091 & 164 & 309 & 18 / 20 / 2 & 26 / 28 / 3 \\
       & Pooled \textsc{gap\_positive} & 0.865 & 0.889 & 0.073 & 0.026 & 0.126 & 0.177 & 369 & 455 & 29 / 31 / 4 & 35 / 37 / 5 \\
      \bottomrule
  \end{tabular}%
  }
\end{table}

\begin{table}[h]
  \centering
  \caption{Axis-effect ranks on $\mathcal{G}$ per task, without (Baseline) and with the oracle tool, for Qwen-122B-A10B and Step-3.7-Flash (1 = largest effect). Unlike \cref{tab:model-family-axis-effects}, the Qwen-122B baseline includes the \emph{act-only} level, so its ranks differ from that table.}
  \label{tab:oracle-axis-effects}
  \small
  \resizebox{\textwidth}{!}{%
  \begin{tabular}{@{}llcccccccc@{}}
      \toprule
      & & \multicolumn{2}{c}{\texttt{promoter}} & \multicolumn{2}{c}{\texttt{redshift}} & \multicolumn{2}{c}{\texttt{rna}} & \multicolumn{2}{c}{Mean Rank} \\
      \cmidrule(lr){3-4} \cmidrule(lr){5-6} \cmidrule(lr){7-8} \cmidrule(lr){9-10}
      Model & Axis & Baseline & Oracle & Baseline & Oracle & Baseline & Oracle & Baseline & Oracle \\
      \midrule
      \multirow{4}{*}{Qwen-122B}
        & Information  & 1 & 1 & 1 & 1 & 1 & 1 & 1.00 & 1.00 \\
        & Reasoning    & 2 & 2 & 3 & 2 & 2 & 2 & 2.33 & 2.00 \\
        & Verification & 3 & 4 & 4 & 4 & 3 & 3 & 3.33 & 3.67 \\
        & Budget       & 4 & 3 & 2 & 3 & 4 & 4 & 3.33 & 3.33 \\
      \addlinespace
      \multirow{4}{*}{Step-3.7-Flash}
        & Information  & 1 & 2 & 1 & 1 & 1 & 1 & 1.00 & 1.33 \\
        & Reasoning    & 3 & 3 & 2 & 3 & 2 & 4 & 2.33 & 3.33 \\
        & Verification & 4 & 4 & 3 & 2 & 3 & 3 & 3.33 & 3.00 \\
        & Budget       & 2 & 1 & 4 & 4 & 4 & 2 & 3.33 & 2.33 \\
      \bottomrule
  \end{tabular}%
  }
\end{table}

\begin{table}[p]
  \centering
  \caption{Axis effects on $\mathcal{G}$ for two open-weight models of similar size: Qwen-122B-A10B and Step-3.7-Flash (198B-A11B). Step-3.7-Flash does not support the \emph{act-only} level of \emph{Reasoning}, so both blocks exclude this level; \emph{Reasoning}'s rank and effect are computed over its remaining two levels.}
  \label{tab:model-family-axis-effects}
  \small
  \begin{tabular}{llccccccc}
    \toprule
    & & \multicolumn{2}{c}{\texttt{promoter}} & \multicolumn{2}{c}{\texttt{redshift}} & \multicolumn{2}{c}{\texttt{rna}} & \\
    \cmidrule(lr){3-4}\cmidrule(lr){5-6}\cmidrule(lr){7-8}
    Model & Axis & Rank & Effect & Rank & Effect & Rank & Effect & Mean Rank \\
    \midrule
    \multirow{4}{*}{Qwen-122B}
      & Information  & 1 & 2.172 & 1 & 2.027 & 1 & 1.410 & 1.00 \\
      & Verification & 2 & 0.256 & 3 & 0.195 & 3 & 0.193 & 2.67 \\
      & Budget       & 4 & 0.086 & 2 & 0.313 & 2 & 0.262 & 2.67 \\
      & Reasoning    & 3 & 0.116 & 4 & 0.028 & 4 & 0.041 & 3.67 \\
    \addlinespace
    \multirow{4}{*}{Step-3.7-Flash}
      & Information  & 1 & 0.848 & 1 & 0.951 & 1 & 1.412 & 1.00 \\
      & Reasoning    & 3 & 0.204 & 2 & 0.298 & 2 & 0.425 & 2.33 \\
      & Verification & 4 & 0.173 & 3 & 0.282 & 3 & 0.247 & 3.33 \\
      & Budget       & 2 & 0.558 & 4 & 0.177 & 4 & 0.058 & 3.33 \\
    \bottomrule
  \end{tabular}
\end{table}

\begin{table}[p]
  \centering
  \caption{Qwen-122B-A10B vs.\ Step-3.7-Flash: outcome, calibration error, cost, and effort, excluding \emph{act-only} as in \cref{tab:model-family-axis-effects}.}
  \label{tab:model-family-cost}
  \small
  \resizebox{\textwidth}{!}{%
  \begin{tabular}{@{}llrrrrr@{}}
      \toprule
      Model & Task(s) & Outcome & Cal. Err. & Cost (USD) & Wall-clock (s) & Steps / TC / Err. \\
      \midrule
      \multirow{2}{*}{Qwen-122B} & \texttt{mmlu-astronomy}$^\dagger$ & 0.702 & 0.153 & 0.089 & 164 & 17.8 / 21.6 / 2.1 \\
       & Pooled \textsc{gap\_positive} & 0.802 & 0.105 & 0.228 & 347 & 25.7 / 31.9 / 3.7 \\
      \addlinespace
      \multirow{2}{*}{Step-3.7-Flash} & \texttt{mmlu-astronomy}$^\dagger$ & 0.677 & 0.285 & 0.043 & 164 & 17.8 / 19.9 / 2.1 \\
       & Pooled \textsc{gap\_positive} & 0.865 & 0.073 & 0.126 & 369 & 28.9 / 30.8 / 4.1 \\
      \bottomrule
  \end{tabular}%
  }
\end{table}

\begin{table}[p]
  \centering
  \caption{Axis effects on $\mathcal{G}$ for an open-weight and a closed model: Qwen-397B-A17B vs.\ Claude Sonnet~5.}
  \label{tab:model-family-l-axis-effects}
  \small
  \begin{tabular}{llccccccc}
    \toprule
    & & \multicolumn{2}{c}{\texttt{promoter}} & \multicolumn{2}{c}{\texttt{redshift}} & \multicolumn{2}{c}{\texttt{rna}} & \\
    \cmidrule(lr){3-4}\cmidrule(lr){5-6}\cmidrule(lr){7-8}
    Model & Axis & Rank & Effect & Rank & Effect & Rank & Effect & Mean Rank \\
    \midrule
    \multirow{4}{*}{Qwen-397B}
      & Information  & 1 & 1.884 & 1 & 1.126 & 1 & 1.051 & 1.00 \\
      & Reasoning    & 3 & 0.467 & 2 & 0.238 & 2 & 0.233 & 2.33 \\
      & Budget       & 2 & 0.497 & 4 & 0.106 & 4 & 0.070 & 3.33 \\
      & Verification & 4 & 0.235 & 3 & 0.207 & 3 & 0.178 & 3.33 \\
    \addlinespace
    \multirow{4}{*}{Claude Sonnet~5}
      & Budget       & 1 & 1.589 & 1 & 1.878 & 1 & 1.796 & 1.00 \\
      & Information  & 2 & 0.521 & 2 & 1.293 & 2 & 1.683 & 2.00 \\
      & Reasoning    & 3 & 0.287 & 3 & 0.370 & 4 & 0.219 & 3.33 \\
      & Verification & 4 & 0.099 & 4 & 0.153 & 3 & 0.279 & 3.67 \\
    \bottomrule
  \end{tabular}
\end{table}

\begin{table}[p]
  \centering
  \caption{Qwen-397B-A17B vs.\ Claude Sonnet~5: outcome, calibration error, cost, and effort.}
  \label{tab:model-family-l-cost}
  \small
  \resizebox{\textwidth}{!}{%
  \begin{tabular}{@{}llrrrrr@{}}
      \toprule
      Model & Task(s) & Outcome & Cal. Err. & Cost (USD) & Wall-clock (s) & Steps / TC / Err. \\
      \midrule
      \multirow{2}{*}{Qwen-397B} & \texttt{mmlu-astronomy}$^\dagger$ & 0.715 & 0.131 & 0.141 & 159 & 19.9 / 25.0 / 2.8 \\
       & Pooled \textsc{gap\_positive} & 0.870 & 0.092 & 0.315 & 342 & 24.7 / 31.4 / 2.6 \\
      \addlinespace
      \multirow{2}{*}{Claude Sonnet~5} & \texttt{mmlu-astronomy}$^\dagger$ & 0.679 & $-$0.043 & 0.220 & 210 & 11.5 / 11.6 / 0.3 \\
       & Pooled \textsc{gap\_positive} & 0.828 & $-$0.016 & 0.407 & 244 & 13.5 / 14.2 / 0.5 \\
      \bottomrule
  \end{tabular}%
  }
\end{table}

\subsection{Medical Tasks}
\label{sec:apx-medical}
This subsection describes the two medical tasks of \cref{sec:medical}. Both have the same structure as the existing tasks: the agent receives a task description, the data, and the required output format, and the sandbox contains six specialist models (the four of the main tasks plus ECGFounder and BioMistral).

\textbf{\texttt{ecg-diagnosis}.} The agent classifies 300 12-lead ECGs from PTB-XL \citep{wagner2020ptb}. The specialist is ECGFounder \citep{li2025electrocardiogram}, a foundation model trained on over 10 million ECGs. Automatic ECG reading helps physicians detect heart rhythm and conduction disorders. Performance is measured by accuracy. ECGFounder reaches $\mathcal{R}=0.950$ zero-shot, and the backbone reaches $\mathcal{B}=0.387$ on rendered ECG plots, so the task is \textsc{gap\_positive}.

\textbf{\texttt{medqa}.} The agent answers 150 four-choice questions from MedQA \citep{jin2021disease}, which are taken from US medical licensing exams and are a standard test of the medical knowledge of language models. The specialist is BioMistral-7B \citep{labrak2024biomistral}, a published medical language model built for such questions. It is therefore a natural model to try, but it reaches only $\mathcal{R}=0.420$, while the backbone alone reaches $\mathcal{B}=0.940$. Like \texttt{mmlu-astronomy}, the task is \textsc{gap\_negative}, and the correct behavior is to decline the specialist. Performance is measured by accuracy.

Instead of the full grid, we vary the three axes behind our main findings: \emph{Information} (4 levels), \emph{Budget} (3 levels), and \emph{Model} (the three Qwen models), with five runs per cell, which gives 180 runs per task. \emph{Reasoning} is fixed to \emph{ReAct} and \emph{Verification} to \emph{asked}. For a fair comparison, the rows for the existing tasks in \cref{tab:medical-slice} are recomputed on the same subset of their main grids. These runs serve as an additional check of our findings. On \texttt{medqa}, all runs (at every information level and with every model) used BioMistral to answer the questions and used the answers without corrections. No run came close to what the agent’s backbone LLM is capable of on its own: the best valid run scored 0.49, against 0.94 for the backbone. This confirms the observation from \texttt{mmlu-astronomy} in a second domain.
\begin{table}[h]
  \centering
  \caption{Medical tasks compared with the same subset of the four existing tasks (\emph{Reasoning} = \emph{ReAct}, \emph{Verification} = \emph{asked}; 180 runs per task). Effect sizes are defined as in \cref{tab:axis-effects}; for the \textsc{gap\_negative} tasks, they are computed on accuracy, as in \cref{sec:apx-mmlu}. Mixed completion is the share of cells in which some but not all runs complete.}
  \label{tab:medical-slice}
  \small
  \resizebox{\textwidth}{!}{%
  \begin{tabular}{@{}llllr@{}}
      \toprule
      & \multicolumn{3}{c}{Effect size} & \\
      \cmidrule(lr){2-4}
      Task(s) & \emph{Information} & \emph{Model} & \emph{Budget} & Mixed completion \\
      \midrule
      Existing four tasks, mean (range) & 1.58 (0.98--2.15) & 0.91 (0.25--1.58) & 0.30 (0.18--0.56) & 42\% (19--72\%) \\
      \texttt{ecg-diagnosis} & 1.08 & 0.81 & 0.11 & 14\% \\
      \texttt{medqa} & 1.39 & 0.90 & 0.06 & 8\% \\
      \bottomrule
  \end{tabular}%
  }
\end{table}

\subsection{Uncertainty of the Axis Effects}
\label{sec:apx-effect-sizes}
The effect size in \cref{tab:axis-effects} carries no uncertainty on its own. We complement it in two ways. First, we bootstrap the ranking: in each of 1{,}000 draws, we resample the runs within every cell with replacement (runs that do not complete stay incomplete), recompute all effect sizes, and record which axis ranks first. Second, we fit a linear model to $\mathcal{G}$ of the completed runs of each task, with all five axes as categorical factors (and, for the interactions, all two-way interactions). We report partial $\eta^2$, the share of the variance that an axis explains after accounting for the other axes, with 95\% intervals from 200 bootstrap draws.
\emph{Information} ranks first on every task in at least 99\% of the draws, and partial $\eta^2$ gives the same order as the effect size, except that \emph{Budget} drops behind \emph{Reasoning}, \emph{Model}, and \emph{Verification} on \texttt{redshift-estimation}. Among the two-way interactions, \emph{Information}$\times$\emph{Model} is the largest on \texttt{promoter-prediction} ($F=12.4$, $p<10^{-12}$, partial $\eta^2=0.04$) and \texttt{rna-folding} ($F=32.1$, $p<10^{-35}$, partial $\eta^2=0.13$). \emph{Budget}$\times$\emph{Model} is significant on \texttt{rna-folding} ($p<10^{-3}$) but not on the other two tasks ($p=0.08$ and $p=0.36$), and \emph{Information}$\times$\emph{Budget} is significant on \texttt{promoter-prediction} ($p=0.002$), borderline on \texttt{redshift-estimation} ($p=0.055$) and not significant on \texttt{rna-folding} ($p=0.18$).
\begin{table}[h]
  \centering
  \caption{Rank stability and partial $\eta^2$ of each axis per task. P(1): share of bootstrap draws in which the axis has the largest effect size. $\eta^2$: partial $\eta^2$ in a main-effects model, with 95\% bootstrap interval.}
  \label{tab:effect-sizes-ci}
  \small
  \resizebox{\textwidth}{!}{%
  \begin{tabular}{@{}lcccccc@{}}
    \toprule
    & \multicolumn{2}{c}{\texttt{promoter-prediction}} & \multicolumn{2}{c}{\texttt{redshift-estimation}} & \multicolumn{2}{c}{\texttt{rna-folding}} \\
    \cmidrule(lr){2-3}\cmidrule(lr){4-5}\cmidrule(lr){6-7}
    Axis & P(1) & $\eta^2$ & P(1) & $\eta^2$ & P(1) & $\eta^2$ \\
    \midrule
    Information  & 1.00 & 0.413 [0.378, 0.445] & 1.00 & 0.296 [0.281, 0.323] & 1.00 & 0.260 [0.231, 0.289] \\
    Model        & 0.00 & 0.080 [0.061, 0.100] & 0.00 & 0.007 [0.002, 0.016] & 0.00 & 0.145 [0.117, 0.177] \\
    Reasoning    & 0.00 & 0.011 [0.004, 0.019] & 0.00 & 0.008 [0.002, 0.017] & 0.00 & 0.008 [0.003, 0.018] \\
    Budget       & 0.00 & 0.009 [0.004, 0.017] & 0.00 & 0.000 [0.000, 0.004] & 0.00 & 0.004 [0.000, 0.012] \\
    Verification & 0.00 & 0.002 [0.000, 0.008] & 0.00 & 0.003 [0.001, 0.012] & 0.00 & 0.000 [0.000, 0.006] \\
    \bottomrule
  \end{tabular}%
  }
\end{table}

\subsection{Interaction Effects and Non-Completion}
\label{sec:apx-completion-robust}
Scores are observed only for runs that complete, and completion itself depends on the \emph{Budget}. A drop in the mean score with more budget could therefore come from weaker runs that now complete, rather than from runs becoming worse. We test this in two ways. First, we restrict the comparison to \emph{matched} configurations, i.e., settings of all axes except \emph{Budget} that produce at least $k$ completed runs under every budget level. Second, we report medians next to means. Confidence intervals come from a bootstrap that resamples runs within each cell (1{,}000 draws). The interactions in \cref{sec:results} largely hold under these checks (\cref{tab:completion-robust}). For the small model (Qwen-35B), the mean and median $\mathcal{G}$ still fall with more budget in matched configurations, and for the large model (Qwen-397B) they rise. The drop for the small model disappears only under the strictest matching (at least three completed runs per budget level: mean $\mathcal{G}$ of 0.81 at \emph{short} and 0.82 at \emph{long}), so part of it is a selection effect. At the \emph{interface} level of \emph{Information}, more budget raises both mean and median. At \emph{none} and \emph{identity}, the mean does not change; at \emph{none}, the median rises slightly.
\begin{table}[h]
  \centering
  \caption{Change in $\mathcal{G}$ from the \emph{short} to the \emph{long} budget (long minus short), pooled over the three \textsc{gap\_positive} tasks, with 95\% bootstrap intervals. \emph{Matched} keeps only configurations with at least one completed run under every budget.}
  \label{tab:completion-robust}
  \small
  \resizebox{\textwidth}{!}{%
  \begin{tabular}{@{}lcccc@{}}
    \toprule
    & \multicolumn{2}{c}{All completed runs} & \multicolumn{2}{c}{Matched configurations ($k=1$)} \\
    \cmidrule(lr){2-3}\cmidrule(lr){4-5}
    Level & Mean & Median & Mean & Median \\
    \midrule
    \emph{Model} = Qwen-35B  & $-0.067$ [$-0.093$, $-0.039$] & $-0.137$ [$-0.185$, $-0.073$] & $-0.045$ [$-0.074$, $-0.017$] & $-0.107$ [$-0.157$, $-0.027$] \\
    \emph{Model} = Qwen-122B & $-0.030$ [$-0.053$, $-0.010$] & $-0.018$ [$-0.033$, $0.000$] & $-0.021$ [$-0.042$, $0.001$] & $-0.004$ [$-0.029$, $0.005$] \\
    \emph{Model} = Qwen-397B & $0.025$ [$0.000$, $0.047$] & $0.015$ [$-0.002$, $0.027$] & $0.036$ [$0.014$, $0.059$] & $0.016$ [$0.004$, $0.028$] \\
    \addlinespace
    \emph{Information} = none      & $-0.002$ [$-0.036$, $0.031$] & $0.057$ [$0.028$, $0.060$] & $0.004$ [$-0.027$, $0.042$] & $0.057$ [$0.027$, $0.060$] \\
    \emph{Information} = identity  & $0.004$ [$-0.035$, $0.040$] & $0.009$ [$0.000$, $0.078$] & $0.020$ [$-0.017$, $0.053$] & $0.012$ [$0.004$, $0.086$] \\
    \emph{Information} = interface & $0.045$ [$0.018$, $0.078$] & $0.028$ [$0.006$, $0.041$] & $0.045$ [$0.014$, $0.076$] & $0.031$ [$0.007$, $0.041$] \\
    \emph{Information} = protocol  & $-0.006$ [$-0.015$, $0.003$] & $0.000$ [$0.000$, $0.000$] & $-0.006$ [$-0.015$, $0.003$] & $0.000$ [$0.000$, $0.000$] \\
    \bottomrule
  \end{tabular}%
  }
\end{table}

\subsection{Generation Speed and Tool Time}
\label{sec:apx-speed}
Because the \emph{Budget} is wall-clock time, a faster model gets more done within the same budget. \Cref{tab:speed} reports, per model, how the wall-clock time of a run splits into LLM generation and tool execution, how many steps a model takes per minute, and how fast it generates. We derive these numbers from the time stamps in the trajectories: the time before an LLM response is counted as generation time, and the time before a tool result as tool time.
Tool execution, not generation, takes most of the wall-clock time for every model (54--73\%), so the budget mostly limits how much code the agent can run. The three Qwen models take a similar number of steps per minute (6.0--7.2), because the smaller model's faster generation is spent on more and shorter steps; between the \emph{short} and \emph{long} budget, the median number of steps per run grows from 25 to 36 for Qwen-35B and from 18 to 24 for the two larger models. 
Claude Sonnet~5 takes fewer steps per minute than the open models and spends a larger share of its time executing code, so its stronger sensitivity to the budget is not explained by faster inference. The speed of generation varied only moderately across runs (CV 0.16--0.26), both on our shared model server and through the Anthropic API.
\begin{table}[h]
  \centering
  \caption{Generation speed and use of time per model, medians over all runs of the respective grid. Steps are LLM calls; LLM s / step is the generation time per step. CV is the coefficient of variation of the generation speed across runs.}
  \label{tab:speed}
  \small
  \resizebox{\textwidth}{!}{%
  \begin{tabular}{@{}lrrrrrr@{}}
    \toprule
    Model & LLM s / step & Steps / min & Tokens / s & CV & LLM share & Tool share \\
    \midrule
    Qwen-35B       & 2.4 & 7.2 & 127 & 0.26 & 30\% & 70\% \\
    Qwen-122B      & 4.1 & 6.0 & 83  & 0.16 & 41\% & 59\% \\
    Qwen-397B      & 4.5 & 6.1 & 70  & 0.24 & 45\% & 55\% \\
    Step-3.7-Flash & 4.3 & 6.0 & 94  & 0.18 & 46\% & 54\% \\
    Claude Sonnet~5 & 4.6 & 3.6 & 84 & 0.16 & 27\% & 73\% \\
    \bottomrule
  \end{tabular}%
  }
\end{table}

\subsection{Use of the Oracle Tool}
\label{sec:apx-oracle-usage}
Since the oracle scores a submission against the test labels, an agent could use it to search for the correct answer instead of verifying its approach. \Cref{tab:oracle-usage} shows how often agents called it. Most runs call it rarely, but 8--22\% of runs call it five times or more. We flag a run as likely hill-climbing when it calls the oracle at least five times and at least half of the repeated calls follow only file edits, without re-running any code in between. This applies to 4.4\% of all oracle runs (210 of 4{,}800), mostly on \texttt{redshift-estimation} (10--15\% of its runs), where the most extreme runs call the oracle up to 111 times while tuning the regression step. On \texttt{mmlu-astronomy}, where changing single answers would be the most direct form of search, we find almost no such runs (0.1--1.5\%). Excluding flagged runs barely changes the pooled $\mathcal{G}$ with the oracle (Qwen-122B: 0.923 to 0.922; Step-3.7-Flash: 0.889 to 0.878), which stays above the baseline without the oracle (0.822 and 0.865).
\begin{table}[h]
  \centering
  \caption{Number of \texttt{oracle\_check} calls per run and share of runs flagged as likely hill-climbing.}
  \label{tab:oracle-usage}
  \small
  \resizebox{\textwidth}{!}{%
  \begin{tabular}{@{}llrrrrrrr@{}}
    \toprule
    Model & Task & 0 & 1 & 2 & 3--4 & $\geq$5 & Mean calls & Flagged \\
    \midrule
    \multirow{4}{*}{Qwen-122B}
      & \texttt{redshift-estimation} & 19.7\% & 38.9\% & 8.3\%  & 11.7\% & 21.4\% & 4.6 & 9.7\% \\
      & \texttt{mmlu-astronomy}      & 8.1\%  & 68.3\% & 8.3\%  & 7.5\%  & 7.8\%  & 1.7 & 0.1\% \\
      & \texttt{promoter-prediction} & 3.8\%  & 44.9\% & 11.0\% & 18.9\% & 21.5\% & 3.0 & 2.5\% \\
      & \texttt{rna-folding}         & 27.8\% & 44.6\% & 9.4\%  & 9.2\%  & 9.0\%  & 1.7 & 2.8\% \\
    \addlinespace
    \multirow{4}{*}{Step-3.7-Flash}
      & \texttt{redshift-estimation} & 44.6\% & 25.8\% & 4.8\%  & 5.6\%  & 19.2\% & 4.0 & 14.6\% \\
      & \texttt{mmlu-astronomy}      & 34.6\% & 25.2\% & 12.3\% & 14.6\% & 13.3\% & 2.0 & 1.5\% \\
      & \texttt{promoter-prediction} & 37.5\% & 24.0\% & 7.9\%  & 11.2\% & 19.4\% & 2.2 & 1.0\% \\
      & \texttt{rna-folding}         & 49.4\% & 27.9\% & 7.9\%  & 5.4\%  & 9.4\%  & 1.9 & 4.0\% \\
    \bottomrule
  \end{tabular}%
  }
\end{table}

\subsection{Axis Analysis for \texttt{mmlu-astronomy}}
\label{sec:apx-mmlu}
\begin{wraptable}{r}{0.5\textwidth}
  \centering
  \caption{Completion rate and mean accuracy per level of \emph{Information} and \emph{Model}. Backbone alone: 0.967; AstroSage reference: 0.671; chance: 0.25.}
  \label{tab:mmlu-axes}
  \small
  \begin{tabular}{@{}llrr@{}}
    \toprule
    Axis & Level & Completion & Acc. \\
    \midrule
    \multirow{4}{*}{\emph{Information}} & none      & 92.8\% & 0.699 \\
                                        & identity  & 93.9\% & 0.712 \\
                                        & interface & 98.7\% & 0.710 \\
                                        & protocol  & 95.7\% & 0.617 \\
    \addlinespace
    \multirow{3}{*}{\emph{Model}} & Qwen-35B  & 89.6\% & 0.626 \\
                                  & Qwen-122B & 97.5\% & 0.707 \\
                                  & Qwen-397B & 98.8\% & 0.715 \\
    \bottomrule
  \end{tabular}
\end{wraptable}
$\mathcal{G}$ is not comparable between the two regimes, so we analyze \texttt{mmlu-astronomy} on raw accuracy (\cref{tab:mmlu-axes}). 
The correct behavior on this task is to answer with the backbone and decline AstroSage, but agents almost never do this. The model-usage heuristic in \cref{tab:model-usage} counts only commands that name the model's path and finds that AstroSage was executed in 66\% of runs; if we also count scripts that load AstroSage and are then executed, 99.8\% of all runs used it. This explains why accuracy (0.685 on average) stays close to AstroSage's own reference accuracy (0.671) and far below the backbone alone (0.967).
Ranked by the same effect size as in \cref{tab:axis-effects}, \emph{Information} (0.93) and \emph{Model} (0.87) matter most, followed by \emph{Budget} (0.15), \emph{Reasoning} (0.08) and \emph{Verification} (0.05). The \emph{protocol} level, which describes how to run AstroSage, lowers accuracy from about 0.71 to 0.62, presumably because it describes the weaker specialist in the most detail and thus points the agent further toward it.

\subsection{Sensitivity to the Hurdle Margin}
\label{sec:apx-margin}
The margin $\mu$ decides when a run counts as clearing the hurdle. We set $\mu$ per task from the sampling noise of its metric. \Cref{tab:margin} repeats the hurdle statistics with $\mu$ set at 0, 0.5, 1, and 2 times its default value per task. The results change little, except in two places. With $\mu=0$ on \texttt{rna-folding}, almost every run clears the hurdle, because its trivial score is 0 and nearly every completed run scores above 0. On \texttt{redshift-estimation}, doubling $\mu$ (a threshold of $R^2>0.5$) removes the weakest hurdle-clearing runs, which raises its ICC. The pooled ICC stays between 45\% and 47\% for every value.
\begin{table}[h]
  \centering
  \caption{Hurdle failure rate and ICC among hurdle-clearing runs for different multiples of the default margin $\mu$. Pooled: three \textsc{gap\_positive} tasks, task means removed.}
  \label{tab:margin}
  \small
  \resizebox{0.77\textwidth}{!}{%
  \begin{tabular}{@{}lcccccccc@{}}
    \toprule
    & \multicolumn{4}{c}{Hurdle failure rate} & \multicolumn{4}{c}{ICC, hurdle-clearing} \\
    \cmidrule(lr){2-5}\cmidrule(lr){6-9}
    Task & $0\mu$ & $0.5\mu$ & $1\mu$ & $2\mu$ & $0\mu$ & $0.5\mu$ & $1\mu$ & $2\mu$ \\
    \midrule
    \texttt{redshift-estimation} & 48.7\% & 53.4\% & 53.8\% & 54.1\% & 35.8\% & 36.2\% & 36.2\% & 53.3\% \\
    \texttt{mmlu-astronomy}      & 4.2\% & 4.3\% & 4.7\% & 5.5\% & 6.9\% & 7.1\% & 6.9\% & 7.2\% \\
    \texttt{promoter-prediction} & 0.2\% & 0.3\% & 0.5\% & 0.6\% & 48.7\% & 48.7\% & 49.7\% & 50.0\% \\
    \texttt{rna-folding}         & 0.2\% & 7.5\% & 8.7\% & 9.5\% & 45.8\% & 35.7\% & 35.2\% & 34.2\% \\
    \midrule
    Pooled                       & 16.4\% & 20.0\% & 20.6\% & 20.9\% & 47.4\% & 44.8\% & 46.1\% & 46.4\% \\
    \bottomrule
  \end{tabular}%
  }
\end{table}

\subsection{Taxonomy Results}
\label{sec:apx-taxonomy}
\begin{figure}[h]
    \centering
    \includegraphics[width=\linewidth]{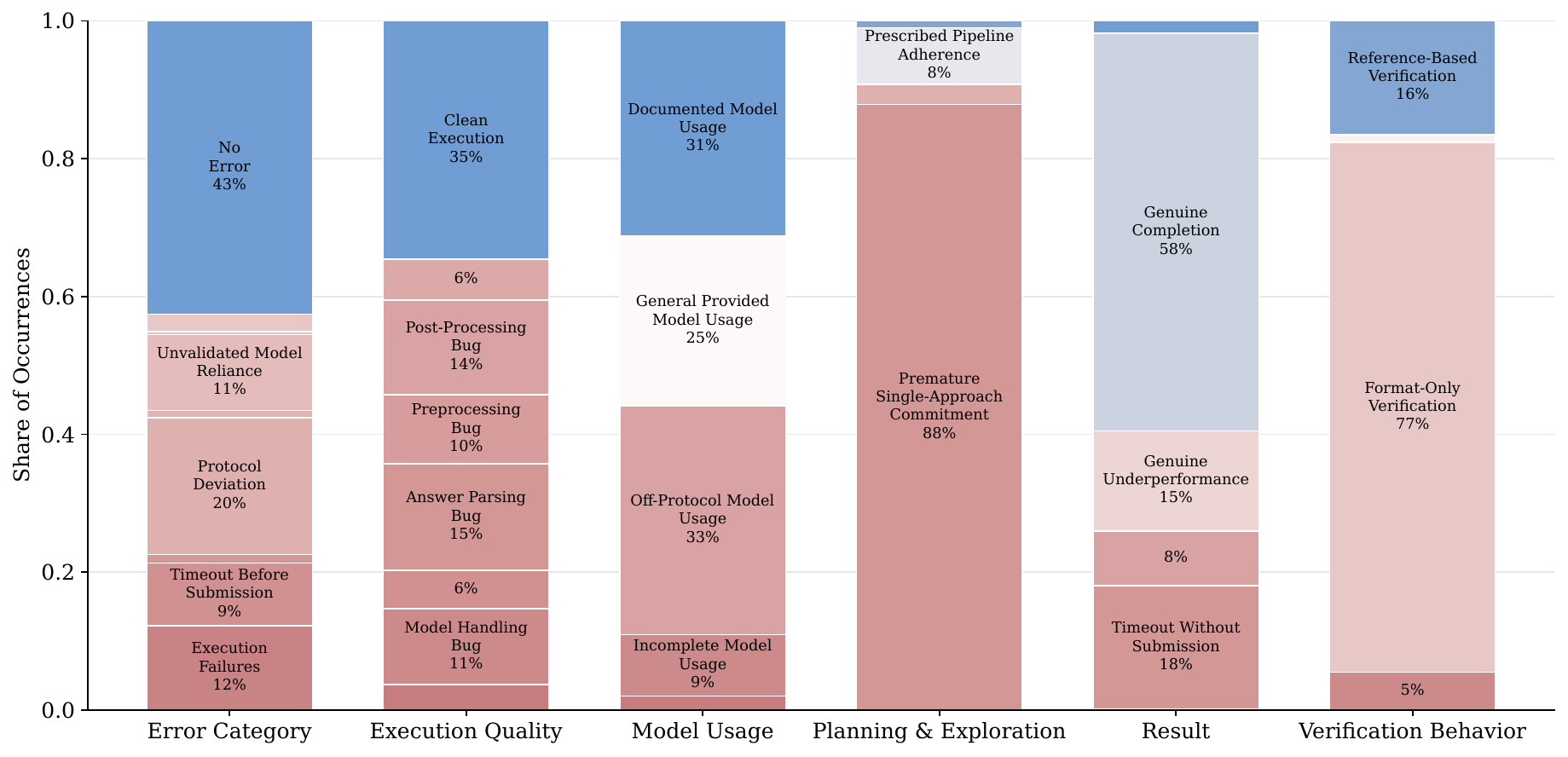}
    \caption{Distribution of categories for each taxonomy dimension in the main experiments.}
    \label{fig:taxonomy-category-distribution}
\end{figure}

\Cref{tab:taxonomy-association} shows how strongly each axis is associated with the category distribution of each taxonomy dimension, measured by Cram\'{e}r's V. \emph{Information} ranks first on all six dimensions, consistent with its dominant effect on scores. \emph{Model} and \emph{Budget} show moderate associations, particularly with error categories and results, while \emph{Reasoning} and \emph{Verification} are only weakly associated with every dimension. \Cref{fig:taxonomy-category-distribution} shows the distribution of categories for each dimension.
\begin{table}[h]
  \centering
  \caption{Association between each axis and the category distribution of each taxonomy dimension in the main experiments, measured by Cram\'{e}r's V (0 = no association, 1 = perfect association). Axes are ranked within each dimension.}
  \label{tab:taxonomy-association}
  \small
  \resizebox{\textwidth}{!}{%
  \begin{tabular}{@{}lcccccccccccc@{}}
    \toprule
    & \multicolumn{2}{c}{Error Category} & \multicolumn{2}{c}{Execution Quality} & \multicolumn{2}{c}{Model Usage} & \multicolumn{2}{c}{Planning \& Exploration} & \multicolumn{2}{c}{Result} & \multicolumn{2}{c}{Verification Behavior} \\
    \cmidrule(lr){2-3} \cmidrule(lr){4-5} \cmidrule(lr){6-7} \cmidrule(lr){8-9} \cmidrule(lr){10-11} \cmidrule(lr){12-13}
    Axis & Rank & V & Rank & V & Rank & V & Rank & V & Rank & V & Rank & V \\
    \midrule
    Information & 1 & 0.233 & 1 & 0.264 & 1 & 0.234 & 1 & 0.271 & 1 & 0.191 & 1 & 0.112 \\
    Reasoning & 5 & 0.041 & 4 & 0.044 & 4 & 0.050 & 4 & 0.038 & 4 & 0.053 & 4 & 0.030 \\
    Verification & 4 & 0.048 & 5 & 0.031 & 5 & 0.020 & 5 & 0.012 & 5 & 0.031 & 5 & 0.023 \\
    Budget & 3 & 0.120 & 3 & 0.070 & 2 & 0.108 & 3 & 0.039 & 2 & 0.152 & 2 & 0.108 \\
    Model & 2 & 0.157 & 2 & 0.138 & 3 & 0.097 & 2 & 0.084 & 3 & 0.095 & 3 & 0.067 \\
    \bottomrule
  \end{tabular}%
  }
\end{table}

\clearpage
\subsection{Agent Prompt}
\label{sec:apx-agent-prompts}
\promptlisting{System Prompt for the Agent}{agent_system_prompt.txt}

\subsection{Specialist Model and Task Examples}
\label{sec:apx-task-examples}
As an example, the following listings show the README of AstroCLIP, the task description of \texttt{redshift-estimation}, and the additional information given at each \emph{Information} level of this task. The \emph{none} level adds no information and therefore has no listing.

\promptlisting{README for \emph{AstroCLIP}}{astroclip_readme.txt}
\promptlisting{Task description for \texttt{redshift-estimation}}{redshift-description.txt}
\promptlisting{Information \emph{identity} for \texttt{redshift-estimation}}{redshift-identity.txt}
\promptlisting{Information \emph{interface} for \texttt{redshift-estimation}}{redshift-interface.txt}
\promptlisting{Information \emph{protocol} for \texttt{redshift-estimation}}{redshift-protocol.txt}

\subsection{Taxonomy Prompts}
\label{sec:apx-taxonomy-prompts}
\promptlisting{System Prompt for the Taxonomy Judge}{taxonomy_system_prompt.txt}
\promptlisting{Prompt for the Taxonomy's \emph{Result} Dimension}{result.txt}
\promptlisting{Prompt for the Taxonomy's \emph{Error Category} Dimension}{error_category.txt}
\promptlisting{Prompt for the Taxonomy's \emph{Specialist Model Usage} Dimension}{model_usage.txt}
\promptlisting{Prompt for the Taxonomy's \emph{Verification Behavior} Dimension}{verification_behavior.txt}
\promptlisting{Prompt for the Taxonomy's \emph{Planning and Exploration} Dimension}{planning_exploration.txt}
\promptlisting{Prompt for the Taxonomy's \emph{Execution Quality} Dimension}{execution_quality.txt}

\end{document}